\documentclass[runningheads]{llncs}

\usepackage{eccv}

\usepackage{eccvabbrv}

\usepackage{graphicx}
\usepackage{booktabs}
\usepackage{wrapfig}
\usepackage[accsupp]{axessibility}  

\definecolor{BestInModule}{RGB}{235,245,255} 
\definecolor{BestOverall}{RGB}{235,255,235}
\definecolor{LightOrange}{RGB}{255, 248, 235}

\usepackage[most]{tcolorbox}
\newtcolorbox[auto counter, number within=section]{namedbox}[2][]{
    colback=white,
    colframe=black,
    fonttitle=\bfseries,
    title=Box~\thetcbcounter: #2,
    #1
}

\usepackage{hyperref}

\usepackage{orcidlink}
\usepackage{color}
\usepackage{multirow}
\usepackage{booktabs}
\usepackage{soul}
\usepackage{colortbl}
\usepackage{bbding}
\usepackage[ruled,linesnumbered]{algorithm2e}
\definecolor{mygray}{gray}{.9}
\definecolor{mypink}{rgb}{.99,.5,.5}
\definecolor{mycyan}{cmyk}{.3,0,0,0}

\usepackage{makecell}

\usepackage{threeparttable}
\usepackage{adjustbox}

\begin{document}

\title{V-Bridge: Bridging Video Generative Priors to Versatile Few-shot Image Restoration} 

\titlerunning{V-Bridge}

\author{Shenghe Zheng\inst{1}\thanks{Equal Contribution. $^{\dagger}$ Corresponding Author.~~Open-source project: \href{https://github.com/Zhengsh123/V-Bridge}{V-Bridge}.~~} \and
Junpeng Jiang\inst{2}$^\star$\and
Wenbo Li\inst{3}$^{\dagger}$}

\authorrunning{Zheng et al.}

\institute{The Hong Kong University of Science and Technology \and
Harbin Institute of Technology, Shenzhen
\and The Chinese University of Hong Kong\\
\email{shenghez.zheng@gmail.com, jjunpeng1122@outlook.com, fenglinglwb@gmail.com}}

\maketitle
\begin{figure*}
\vspace{-0.7cm}
\centering  
\includegraphics[width=0.96\textwidth]{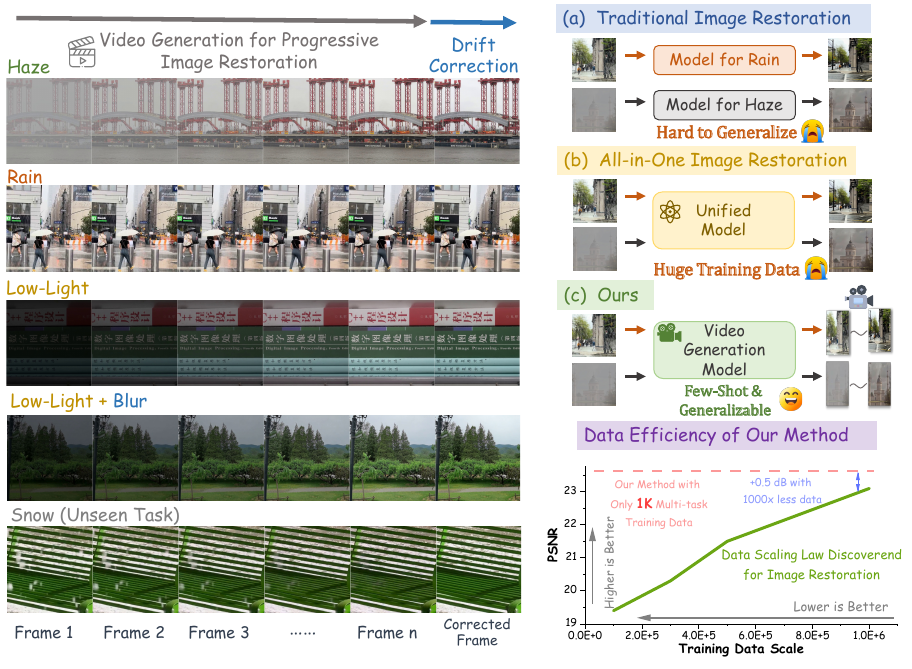} 
\vspace{-0.1cm}
\caption{Left: Image restoration is formulated as progressive video generation with frame drift correction. Right: Leveraging video generative priors leads to stronger generalization under limited data compared to current image restoration method~\cite{foundir}.} 
\label{figure:motivation.}  
\vspace{-20pt}
\end{figure*} 

\begin{abstract}
  Large-scale video generative models are trained on vast and diverse visual data, enabling them to internalize rich structural, semantic, and dynamic priors of the visual world. While these models have demonstrated impressive generative capability, their potential as general-purpose visual learners remains largely untapped. In this work, we introduce V-Bridge, a framework that bridges this latent capacity to versatile few-shot image restoration tasks. We reinterpret image restoration not as a static regression problem, but as a progressive generative process, and leverage video models to simulate the gradual refinement from degraded inputs to high-fidelity outputs. Surprisingly, with only 1,000 multi-task training samples (less than 2\% of existing restoration methods), pretrained video models can be induced to perform competitive image restoration, achieving multiple tasks with a single model, rivaling specialized architectures designed explicitly for this purpose. Our findings reveal that video generative models implicitly learn powerful and transferable restoration priors that can be activated with only extremely limited data, challenging the traditional boundary between generative modeling and low-level vision, and opening a new design paradigm for foundation models in visual tasks.
\end{abstract}

\section{Introduction}
\label{sec:intro}
Large-scale video generative models have recently emerged as powerful visual models~\cite{wan2025, cof, seeddance}. Trained on massive and diverse video corpora, they internalize not only appearance statistics but also structural regularities, object dynamics, lighting variations, and long-range spatio-temporal coherence. Although primarily optimized for video synthesis, the scale, diversity, and structural richness of their training data imply that these models encode far more general visual priors. Such priors extend beyond generation and suggest substantial untapped potential for a wide range of visual understanding and reconstruction tasks.

Among these visual problems, image restoration~\cite{ir} remains largely confined to task-specific modeling. From denoising to deblurring, dominant approaches rely on carefully engineered architectures trained with substantial supervision for each degradation type~\cite{swinir, real-esrgan, autodial}. Despite their efficacy, these paradigms remain decoupled from the rapid advances in generative modeling that have redefined high-level vision. Consequently, they necessitate massive supervision, even exceeding a million samples~\cite{foundir}, to learn restoration from scratch for each degradation. This data-intensive approach underutilizes the rich, transferable priors already embedded within large-scale generative models.

In this work, we reimagine the conventional methodology by recasting image restoration as a video generation process, simulating progressive restoration dynamics instead of performing static, one-step regression. Specifically, the degraded image is treated as the initial state, while the high-fidelity reconstruction serves as the terminal point along a quality-refinement trajectory. This formulation allows extensive video generation priors to be seamlessly and efficiently integrated into image restoration tasks, potentially alleviating the massive data requirements typical of traditional paradigms.

\begin{figure*}[t]
\centering  
\includegraphics[width=1 \textwidth]{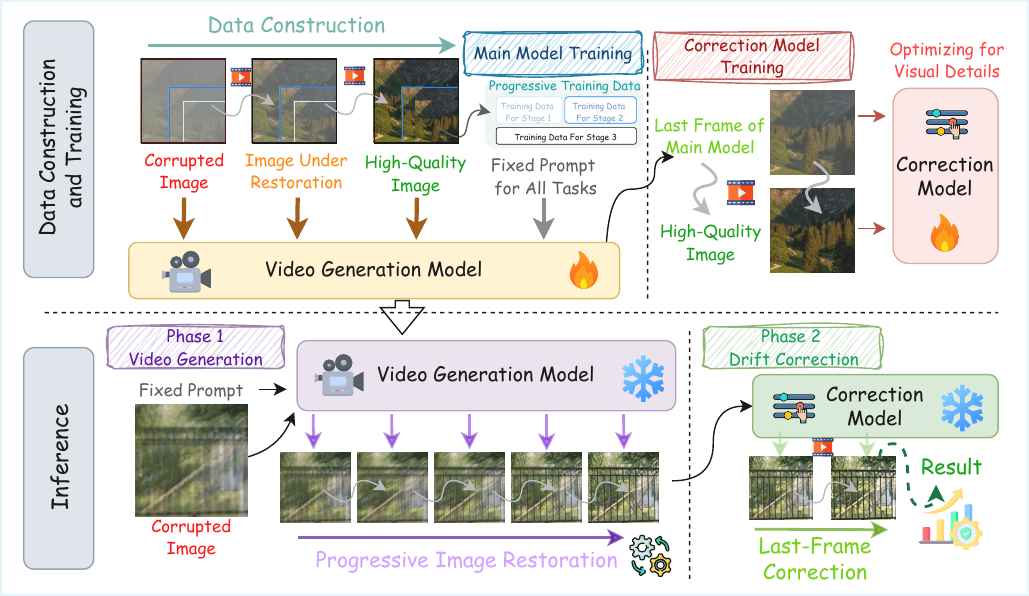} 
\vspace{-0.7cm}
\caption{Overview of the proposed pipeline. The upper part shows data construction and training, where paired low- and high-quality images are used to build pseudo-temporal sequences for progressive restoration learning. A progressive resolution training strategy is adopted to improve fine-grained detail modeling, and an auxiliary generative model is trained for final-frame correction. The lower part shows inference, where the model generates a restoration trajectory and uses the refined last frame as the final output.} 
\label{figure:framework}  
\vspace{-10pt}
\end{figure*}

Driven by this perspective, we introduce V-Bridge (Fig.~\ref{figure:framework}), a framework that harnesses video generative priors for versatile few-shot image restoration, requiring less than 2\% of the training data typical of contemporary methods. V-Bridge models image restoration as a step-wise quality evolution toward high-fidelity outputs. To bridge the resolution gap between moderate-resolution video pretraining and high-resolution restoration, we propose a coarse-to-fine training curriculum that progressively optimizes the model across increasing scales. This strategy allows the model to first establish global structural coherence before refining high-frequency details, thereby ensuring computational efficiency. Furthermore, we incorporate a drift correction module with minimal overhead to enhance fine-grained texture and color fidelity. Extensive experiments demonstrate that V-Bridge transforms a single video generation model into a versatile restoration expert with only 1,000 multi-task training samples. Our approach achieves a 1.6dB gain over baselines trained on 15$\times$ to 1,000$\times$ more data. Remarkably, as shown in Fig.~\ref{figure:motivation.}, V-Bridge generalizes effectively to unseen tasks, showcasing superior out-of-distribution adaptability. Our work opens a new avenue for leveraging rich video priors in low-level vision, paving the way for more general and versatile unified visual models.

Our contributions are threefold:
\begin{itemize}
\item \textbf{New Restoration Paradigm}: We pioneer the use of video generative models as universal priors, demonstrating that their inherent representations serve as a powerful, transferable foundation across diverse low-level tasks.
\item \textbf{The V-Bridge Framework}: We propose V-Bridge, a framework designed for data-efficient image restoration via progressive generative refinement. We introduce a coarse-to-fine training curriculum together with a lightweight drift correction mechanism, enabling sophisticated quality enhancement with minimal task-specific supervision.
\item \textbf{Empirical Validation}: Extensive evaluations reveal that V-Bridge achieves state-of-the-art results with extreme data sparsity (using only 1K samples). Our findings validate the extraordinary out-of-distribution adaptability of video priors and point toward a unified future for low-level modeling.
\end{itemize}
\section{Related Work}
\label{sec:related}

\subsection{Video Generation}
With the success of diffusion models, an increasing number of studies have extended them to video generation. Early approaches typically adopted UNet-based architectures with 2D VAE~\cite{ho2022imagen,blattmann2023stable,chen2023videocrafter1}. However, these designs struggled to achieve substantial performance breakthroughs in terms of scalability and temporal coherence. Inspired by the strong scalability demonstrated by Sora~\cite{videoworldsimulators2024}, the field has shifted toward large-scale video generation models built upon 3D VAE and Diffusion Transformers (DiT)~\cite{dit}. State-of-the-art systems now include a series of open-source models such as OpenSora~\cite{opensora}, HunyuanVideo~\cite{hunyuanvideo}, and Wan~\cite{wan2025}, as well as commercial models including Kling~\cite{kling}, Seedance~\cite{seeddance}, and Veo~\cite{veo}. Notably, recent advances from Veo and Seedance demonstrate remarkable capability in producing temporally consistent and semantically realistic video content. These developments suggest that video generative models are evolving beyond content synthesis, showing strong potential as general visual foundation models for unified representation learning across diverse vision tasks.

\subsection{All-in-One Image Restoration}
Traditional image restoration methods are typically designed for a single predefined degradation type, such as motion blur, rain streaks, or noise, requiring separate models for different corruption scenarios~\cite{jin2023dnf,fang2022robust,guo2022image,liang2021swinir}. While effective within narrow settings, these task-specific designs limit scalability and practical deployment. All-in-one image restoration aims to handle diverse degradations within a single unified model. Early efforts focused on degradation-aware representation learning, where AirNet~\cite{AirNet} utilizes contrastive learning and methods like PromptIR~\cite{prompter} and ProRes~\cite{ma2023prores} introduce lightweight, learnable visual prompt modules to dynamically adapt the network to specific input conditions. To further refine this process, Perceive-IR~\cite{zhang2025perceive} leverages multi-level quality-driven prompt for fine-grained quality control across various degradation types and severity levels. Diffusion-based frameworks such as DiffUIR~\cite{differ} and DiffBIR~\cite{lin2024diffbir} have been introduced to leverage generative priors for higher perceptual quality. Advanced models like AutoDIR~\cite{autodial} and InstructIR~\cite{instructir} further involve the continuous guidance of image restoration via human language instructions However, these approaches still typically require large-scale training and do not fully exploit large-scale visual priors. In contrast, our work explores restoration from the perspective of leveraging pretrained video generative priors as a general visual prior, enabling progressive quality refinement without designing separate restoration models for each degradation type.

\subsection{Chain-of-Frames Reasoning}
The rapid maturation of video generation models has catalyzed a paradigm shift from simple motion synthesis to complex visual inference, a phenomenon encapsulated by the Chain-of-Frames (CoF) reasoning~\cite{cof}. To systematically quantify this emergent intelligence, a diverse array of empirical studies and specialized benchmarks has been established, scrutinizing model performance across dimensions such as spatial relationships, logical reasoning, action planning, and physical dynamics~\cite{deng2025video,guo2025video,luo2025v,yang2025reasoning,li2025viper}. Parallel to these evaluative efforts, recent advancements have focused on augmenting the CoF reasoning capabilities of video models through supervised fine-tuning (SFT) on curated video sequences~\cite{miniveo3reasoner} and test-time prompt optimization~\cite{chen2025tivibench}. More recently, the versatility of CoF reasoning has been extended to text-to-image synthesis, yielding promising results by treating image generation as the final-state of a reasoning chain~\cite{tong2026cof}. However, while existing literature predominantly focuses on high-level semantic and logical orchestration, the potential of CoF-driven temporal priors to address low-level vision tasks remains a conspicuously unmapped frontier. This leaves a significant research gap: the question of whether the structured \textit{visual thinking} inherent in CoF can be harnessed to resolve granular pixel-level challenges or restore fine-grained structural integrity remains entirely unexplored.

\section{Methodology}
\label{sec:method}

\subsection{Overview}
In this section, we present V-Bridge, a progressive restoration framework that repurposes pretrained video generative priors for image-to-image translation. Unlike vanilla one-step regression, we reformulate restoration as a temporally evolving trajectory that iteratively refines image quality. By harnessing the inherent spatio-temporal consistency and generative priors of video generation models, V-Bridge achieves remarkable performance with only 0.1\% to 2\% of the task-specific training data required by current methods~\cite{autodial,foundir}.

The methodology is organized as follows: Sec.~\ref{sec:method:data} details the construction of pseudo-temporal sequences from paired low-quality (LQ) and high-quality (HQ) images. Sec.~\ref{sec:method:training} presents a Progressive Curriculum Training strategy that transitions from structural recovery to fine-grained synthesis. Finally, Sec.~\ref{sec:method:inference} describes a Drift Correction mechanism designed to bridge the resolution gap between video generative priors and high-resolution restoration. Through this dynamic formulation, we transform static restoration into a learnable flow problem, fully unlocking the few-shot potential of video foundation models.



\subsection{Pseudo-Temporal Data Construction}\label{sec:method:data}

To translate static restoration into a dynamic generation task, we we lift each low-quality--high-quality (LQ-HQ) pair 
$(\mathbf{I}_{\mathrm{LQ}}, \mathbf{I}_{\mathrm{HQ}})$ 
into a pseudo-temporal sequence with explicit quality progression.

Given $\mathbf{I}_{\mathrm{LQ}}$ as the anchor (initial frame) and $\mathbf{I}_{\mathrm{HQ}}$ as the target (terminal frame), we construct a sequence $\{\mathbf{I}_t\}_{t=0}^{T}$ of length $T+1$ such that:
\begin{align}
\mathbf{I}_0 = \mathbf{I}_{\mathrm{LQ}}, \quad
\mathbf{I}_T = \mathbf{I}_{\mathrm{HQ}}.
\end{align}

For intermediate frames $t \in \{1, \dots, T-1\}$, we define a continuous transition path in pixel space via linear interpolation:
\begin{align}
\mathbf{I}_t = (1 - \alpha_t)\mathbf{I}_{\mathrm{LQ}} + \alpha_t \mathbf{I}_{\mathrm{HQ}}, \
\alpha_t = \frac{t}{T}.
\end{align}

This formulation encapsulates a monotonic quality evolution, providing temporally consistent supervision that guides the model to learn a stable low-to-high quality trajectory. By converting static pairs into a pseudo-video stream, we provide the video model with temporally consistent supervision, enabling it to learn the entire restoration trajectory rather than a singular mapping.



\subsection{Progressive Curriculum Training}\label{sec:method:training}
In this section, we introduce a multi-stage training strategy for efficient and effective restoration. The overall optimization objective remains identical across all stages and follows a supervised fine-tuning paradigm, where the model learns to regress the constructed progressive data sequences. The stage-wise design does not modify the objective itself, but progressively improves the model’s ability to synthesize fine details and enhance fidelity.

\noindent\textbf{Overall Training Objective.} Formally, let $\{\mathbf{I}_t\}_{t=0}^{T}$ denote the pseudo-temporal sequence and $\theta$ the model parameters. 
Given the conditional input $\mathbf{I}_0$ and the time index $t$, the model predicts $\hat{\mathbf{I}}_t = f_{\theta}(\mathbf{I}_0, t)$. 
The training objective is:
\begin{align}
\mathcal{L}(\theta) 
&= \mathbb{E}_{(\mathbf{I}_0, \mathbf{I}_t)} 
\left[ \, \ell \big( f_{\theta}(\mathbf{I}_0, t), \mathbf{I}_t \big) \right],
\end{align}
where $\ell(\cdot,\cdot)$ denotes a reconstruction loss. This formulation encourages the video model to mimic the image restoration process by progressively approximating the intermediate states, thereby fully unleashing its potential for restoration tasks while learning the complete low-to-high quality trajectory.

\noindent\textbf{Curriculum Training.} A key gap between image restoration and video generation lies in training resolution: existing video generative models are rarely trained on high-resolution data (e.g., 4K), which is crucial for fine-detail restoration. However, directly training at such resolutions is computationally expensive and may reduce learning efficiency due to the pre-training–fine-tuning data gap.

To overcome this, we construct a progressive resolution curriculum $\{r_t\}_{t=1}^T$, which controls the difficulty of restoration learning by modulating spatial fidelity across training stages. Let $\{v_i\}_{i=1}^N$ denote the training corpus, where each video $v_i=\{f_{i,1},\dots,f_{i,T_i}\}$ is a spatio-temporal sample from the video distribution. Instead of training on the original high-resolution data, we apply stage-dependent downsampling. At stage 
$t$, video samples are re-encoded using a resolution-aware degradation operator as following:
\begin{align}
v_i^{(t)} = \mathrm{DownUp}(v_i, r_t), \quad r_t \in \{r_1, r_2, \dots, r_T\}, \; r_1 < r_2 < \cdots < r_T.
\end{align}

In simple terms, the video resolution is gradually increased during training. This enables the model to first capture global restoration at low resolution and then progressively enhance fine-grained detail generation as resolution grows.

From a generative modeling perspective, this progressive learning strategy discretizes a continuous restoration probability flow by learning conditional transition kernels across quality levels. By gradually increasing resolution complexity, the model captures hierarchical semantics and high-frequency perceptual statistics in a coarse-to-fine manner, improving few-shot generalization while preserving perceptual fidelity and temporal consistency.
\subsection{Drift Correction}\label{sec:method:inference}
The proposed curriculum training reduces cost and optimization difficulty but cannot fully bridge the large gap between pretraining resolution and the fine-grained detail generation capability required in our task. Most video generation models are pretrained at moderate resolutions (e.g., $720\text{p}$), whereas practical restoration tasks often require recovering ultra high-resolution content (e.g., $4\text{K}$). This discrepancy limits the model’s ability to faithfully recover high-quality fine details, often leading to struggles in reconstructing high-frequency structures.

We interpret this limitation as an implicit distribution drift induced by the resolution-constrained generative prior. Let $\hat{x}$ denote the final restored frame predicted by the base video generation model, and let $x^{\mathrm{HR}}$ denote the corresponding ground-truth high-resolution image. Due to the pretrained resolution bias, $\hat{x}$ exhibits a systematic drift from the target high-fidelity manifold, and can be viewed as a sample drawn from a lower-fidelity distribution:
\begin{align}
\hat{x} \sim p_{\theta}^{\text{LR}}(x), 
\qquad 
x^{\mathrm{HR}} \sim p_{\text{HR}}(x),
\end{align}
where $p_{\theta}^{\text{LR}}$ represents the distribution distorted by low-resolution pretraining.

To mitigate this drift, we introduce an additional drift correction model that explicitly learns a short corrective trajectory from $\hat{x}$ toward $x^{\mathrm{HR}}$. The training samples are constructed as short pseudo-temporal sequences interpolating between the drifted base model output and the ground-truth high-quality image, enabling a smooth transition from resolution-limited restoration to full-fidelity reconstruction. We view this transformation as a form of degradation type tailored to the unique characteristics of video generative models.

The drift correction model is trained to parameterize this conditional generative transition, effectively modeling a mapping:
\begin{align}
g_{\phi}: \; p_{\theta}^{\text{LR}}(x) \rightarrow p_{\text{HR}}(x),
\end{align}
such that the final output approximates $x^{\mathrm{HR}}$ while preserving structural consistency with $\hat{x}$. By restricting the trajectory to only a few intermediate frames, the correction process remains computationally efficient, yet substantially eliminates the resolution-induced bias and enhances perceptual quality.

\begin{figure*}[t]
\centering  
\includegraphics[width=1 \textwidth]{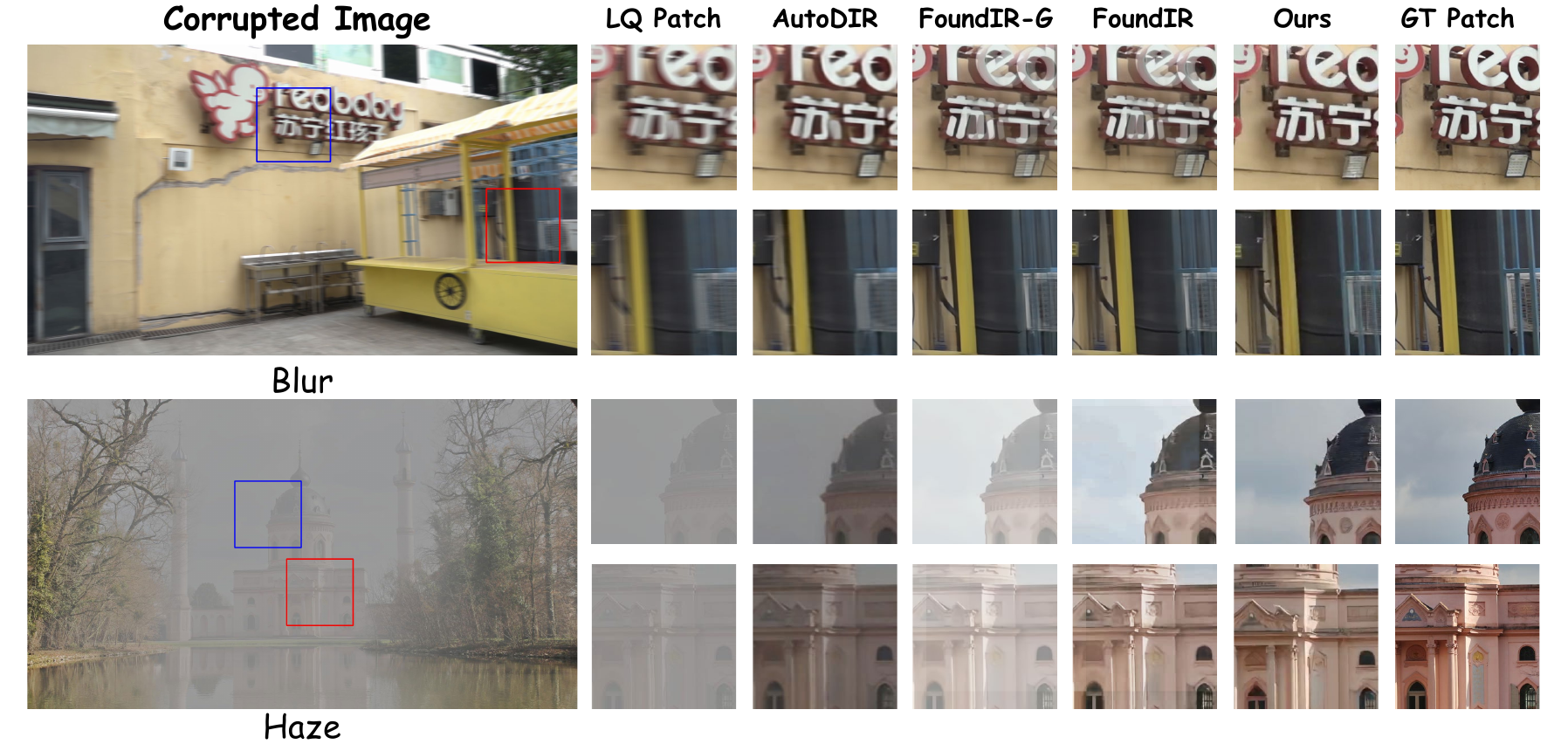} 
\vspace{-0.3cm}
\caption{Visualization results on a subset of the FoundIR test set. FoundIR-G is the generalist model of FoundIR. GT denotes ground truth. Bounding boxes with different colors indicate zoomed regions for detailed comparison. Compared with other methods, our approach achieves higher visual fidelity and stronger structural consistency, while showing superior robustness across diverse degradation patterns.}\label{fig:exp:foundir-visual}
\vspace{-5pt}
\end{figure*}

\section{Experiments}
\label{sec:exp}

\subsection{Experimental Settings}
\noindent\textbf{Training Details.} Our training data are sampled from FoundIR~\cite{foundir} and RealCE~\cite{real-ce}. Progressive training sequences are constructed as described in Sec.~\ref{sec:method:data} to facilitate gradual restoration learning. Unless otherwise specified, we randomly select 50 samples per task category from each dataset (FoundIR and RealCE) for training, and also use 50 samples per category for the drift correction stage. Both models adopt Wan2.2-TI2V-5B~\cite{wan2025} as the backbone network. More implementation and training details are provided in Appendix~\ref{appendix:training-details}.

\noindent\textbf{Test Details.} We evaluate our method on the FoundIR~\cite{foundir} test split, covering diverse degradations including blur, noise, JPEG compression, haze, rain, raindrop, low-light conditions, and mixed degradations. In addition, we report results on several external benchmark datasets~\cite{dense-haze, uhd-ll, nh-haze, uav-rain1k, hq-nightrain,weatherbench} to further validate cross-dataset generalization and real-world applicability. We further evaluate out-of-distribution (OOD) robustness in Sec.~\ref{sec:exp:analysis:ood} to examine model stability under unseen and more severe degradation scenarios. These evaluations are designed to comprehensively verify both restoration quality and generalization capability. More implementation and evaluation details are provided in Appendix~\ref{appendix:test-details}.

\noindent\textbf{Evaluation Metrics.} We evaluate restoration quality using PSNR and SSIM, two widely adopted metrics in image restoration. PSNR quantifies pixel-wise reconstruction fidelity, while SSIM measures structural consistency between the restored image and the ground truth. Higher values for both metrics indicate better performance. Details are provided in the Appendix~\ref{appendix:test-details}.

\begin{table*}[t]
\centering
\caption{Quantitative comparison ($\frac{\mathrm{PSNR}}{\mathrm{SSIM}}$, both higher is better) on the FoundIR test set. FoundIR-G denotes the generalist model of FoundIR and DC denotes Drift Correction. Underlining and boldface denote the second-best and best methods, respectively.}
\label{tab:comparison}
\vspace{-4pt}
\scriptsize
 \begin{adjustbox}{width=\textwidth}
\begin{tabular}{l|cccccccccc|cc}
\toprule
 & \multirowcell{2}{{Real-}\\{ESRGAN}} & \multirow{2}{*}{DGUNet} & \multirowcell{2}{{Trans-}\\{Weather}} & \multirow{2}{*}{PromptIR} & \multirow{2}{*}{DiffUIR} & \multirow{2}{*}{DA-CLIP} & \multirow{2}{*}{X-Restormer} & \multirow{2}{*}{InstructIR} & \multirow{2}{*}{AutoDIR} & \multirow{2}{*}{FoundIR-G} & \multirowcell{2}{{Ours}\\{w/o DC}} & \multirow{2}{*}{Ours} \\
 & &  &  &  &  &  &  &  &  &  &  &  \\\midrule
 
\textbf{Data Scale}
& \cellcolor{BestInModule}\underline{15K} & 16K & 19K & 77K & 30K & 53K & 40K & \cellcolor{BestInModule}\underline{15K} & 23K & 1M & \cellcolor{BestOverall}\textbf{1K} & \cellcolor{BestOverall}\textbf{1K} \\ \midrule

\multirow{2}{*}{Blur}
& \cellcolor{BestInModule}\underline{25.20} & 21.86 & 21.34 & 21.91 & \cellcolor{BestOverall}{\textbf{25.31}} & 20.92 & 21.88 & 20.15 & 20.31 & 24.34 & 21.02 & 24.92 \\

& \cellcolor{BestInModule}\underline{0.7868} & 0.7334 & 0.7103 & 0.7339 & \cellcolor{BestOverall}\textbf{0.7979} & 0.6954 & 0.7325 & 0.6801 & 0.6946 & 0.7856 & 0.6862 & 0.7809 \\
\midrule

\multirow{2}{*}{Noise}
& 34.46 & 36.69 & 30.12 & 36.57 & 34.48 & 29.45 & 35.86 & \cellcolor{BestInModule}\underline{38.58} & 36.84 & \cellcolor{BestOverall}\textbf{38.61} & 32.61 & 32.87 \\
& 0.9585 & 0.9494 & 0.8945 & 0.9478 & 0.9622 & 0.9027 & 0.9412 & \cellcolor{BestInModule}\underline{0.9628} & 0.9221 & \cellcolor{BestOverall}\textbf{0.9662} & 0.9397 & 0.9498 \\
\midrule

\multirow{2}{*}{JPEG}
& 27.62 & 29.80 & 23.52 & 29.63 & 30.09 & 25.77 & 28.58 & \cellcolor{BestInModule}\underline{32.99} & 32.77 & \cellcolor{BestOverall}\textbf{34.03} & 28.21 & 26.59 \\
& 0.9108 & 0.8906 & 0.7296 & 0.8842 & \cellcolor{BestOverall}\textbf{0.9390} & 0.7816 & 0.8646 & 0.9378 & 0.9280 & \cellcolor{BestInModule}\underline{0.9379} & 0.8487 & 0.8269 \\
\midrule

\multirow{2}{*}{Haze}
& \cellcolor{BestOverall}\textbf{22.07} & 18.58 & 17.72 & 15.36 & 19.97 & 15.91 & 16.47 & 16.85 & 15.23 & 16.65 & 21.25 & \cellcolor{BestInModule}\underline{21.70} \\
& \cellcolor{BestOverall}\textbf{0.8380} & 0.6762 & 0.5635 & 0.4982 & \cellcolor{BestInModule}\underline{0.8193} & 0.5399 & 0.6257 & 0.7555 & 0.6264 & 0.7814 & 0.6918 & 0.6711 \\
\midrule

\multirow{2}{*}{Rain}
& 28.95 & 25.47 & 23.49 & 27.59 & 30.17 & 23.04 & 25.70 & \cellcolor{BestInModule}\underline{30.18} & 25.69 & \cellcolor{BestOverall}\textbf{33.09} &26.40 & 25.42 \\
& 0.9226 & 0.8245 & 0.6886 & 0.8411 & \cellcolor{BestInModule}\underline{0.9350} & 0.6849 & 0.8076 & 0.8997 & 0.7711 & \cellcolor{BestOverall}\textbf{0.9387} & 0.7957 & 0.7836 \\
\midrule

\multirow{2}{*}{Raindrop}
& \cellcolor{BestOverall}\textbf{28.94} & 24.83 & 22.94 & 25.83 & 26.98 & 20.86 & 26.11 & 21.05 & 20.82 & \cellcolor{BestInModule}\underline{28.52} & 25.77 & 24.99 \\
& \cellcolor{BestOverall}\textbf{0.9115} & 0.7924 & 0.6814 & 0.8327 & 0.8908 & 0.6490 & 0.8368 & 0.6828 & 0.6687 & \cellcolor{BestInModule}\underline{0.9110} & 0.7698 & 0.7380 \\
\midrule

\multirow{2}{*}{Lowlight}
& 19.26 & 16.16 & 14.95 & 16.33 & 14.02 & 17.34 & 16.02 & 20.04 & \cellcolor{BestInModule}\underline{21.90} & 12.35 & 19.18 & \cellcolor{BestOverall}\textbf{26.94} \\
& \cellcolor{BestInModule}\underline{0.8709} & 0.6494 & 0.6295 & 0.6353 & 0.7101 & 0.7412 & 0.6404 & 0.8542 & 0.8288 & 0.7185 & 0.8278 & \cellcolor{BestOverall}\textbf{0.8944} \\
\midrule

\multirow{2}{*}{B+N}
& 23.48 & 22.90 & 22.19 & 22.93 & \cellcolor{BestInModule}\underline{24.44} & 22.15 & 22.90 & 21.70 & 21.90 & 22.53 & 23.80 & \cellcolor{BestOverall}\textbf{27.31} \\
& 0.7728 & 0.7633 & 0.7363 & 0.7587 & \cellcolor{BestInModule}\underline{0.8039} & 0.7293 & 0.7544 & 0.7185 & 0.7293 & 0.7654 & 0.7703 & \cellcolor{BestOverall}\textbf{0.8471} \\
\midrule

\multirow{2}{*}{B+J}
& 20.41 & 22.86 & 22.14 & 22.90 & 21.36 & 21.36 & 22.85 & 21.39 & 22.03 & \cellcolor{BestOverall}\textbf{28.33} & 24.05 & \cellcolor{BestInModule}\underline{25.33} \\
& 0.6562 & 0.7358 & 0.7105 & 0.7360 & 0.6915 & 0.6940 & 0.7321 & 0.6814 & 0.7088 & \cellcolor{BestOverall}\textbf{0.8491} & 0.7571 & \cellcolor{BestInModule}\underline{0.8020}\\
\midrule

\multirow{2}{*}{N+J}
& 29.71 & 35.28 & 25.59 & 35.08 & 31.96 & 26.03 & 33.83 & \cellcolor{BestOverall}\textbf{39.90} & 37.55 & \cellcolor{BestInModule}\underline{39.21} & 33.33 & 31.63 \\
& 0.9566 & 0.9564 & 0.8377 & 0.9524 & \cellcolor{BestOverall}\textbf{0.9782} & 0.8602 & 0.9405 & \cellcolor{BestInModule}\underline{0.9770} & 0.9616 & 0.9745 & 0.9422 & 0.9323 \\
\midrule

\multirow{2}{*}{R+H}
& \cellcolor{BestOverall}\textbf{20.40} & 18.79 & 18.43 & 16.61 & \cellcolor{BestInModule}\underline{20.24} & 13.97 & 15.17 & 13.49 & 14.90 & 15.42 & 18.70 & 18.69 \\
& \cellcolor{BestInModule}\underline{0.7153} & 0.4981 & 0.4245 & 0.4985 & \cellcolor{BestOverall}\textbf{0.7268} & 0.3563 & 0.4354 & 0.5535 & 0.4213 & 0.6766 & 0.5574 & 0.5222 \\
\midrule

\multirow{2}{*}{L+H}
& \cellcolor{BestOverall}\textbf{21.79} & 13.20 & 15.99 & 15.36 & 19.31 & 12.62 & 14.75 & 13.52 & 14.50 & 17.05 & 19.74 & \cellcolor{BestInModule}\underline{20.79} \\
& \cellcolor{BestOverall}\textbf{0.8084} & 0.4182 & 0.4285 & 0.4982 & 0.7745 & 0.3220 & 0.4416 & 0.4983 & 0.4456 & 0.7556 & 0.5983 & \cellcolor{BestInModule}\underline{0.6209} \\
\midrule

\multirow{2}{*}{L+R}
& 32.16 & 30.66 & 28.29 & 31.03 & 32.35 & 21.58 & 30.93 & 29.87 & 27.15 & \cellcolor{BestInModule}\underline{32.77} &\cellcolor{BestOverall}\textbf{32.90} & 31.27 \\
& 0.9116 & 0.9097 & 0.8663 & 0.9039 & \cellcolor{BestInModule}\underline{0.9255} & 0.7246 & 0.9026 & 0.8866 & 0.8576 & \cellcolor{BestOverall}\textbf{0.9281} &  0.9126 & 0.9064 \\
\midrule

\multirow{2}{*}{L+B}
& \cellcolor{BestInModule}\underline{21.95} & 11.54 & 21.35 & 20.29 & 18.97 & 16.17 & 19.38 & 17.43 & 19.14 & 18.07 & 20.37 & \cellcolor{BestOverall}\textbf{23.78} \\
& 0.7254 & 0.4922 & 0.6764 & 0.6723 & 0.6939 & 0.6241 & 0.6605 & 0.6676 & 0.6570 & \cellcolor{BestOverall}\textbf{0.7551} & 0.6543 & \cellcolor{BestInModule}\underline{0.7378} \\
\midrule

\multirow{2}{*}{L+N}
& 17.49 & 12.56 & 11.56 & 10.83 & 13.88 & 15.70 & 9.75 & 16.37 & \cellcolor{BestInModule}\underline{17.49} & 12.77 & 17.36 & \cellcolor{BestOverall}\textbf{24.60} \\
& 0.7358 & 0.5349 & 0.5181 & 0.4499 & 0.6262 & 0.6365 & 0.3866 & 0.4625 & 0.6900 & \cellcolor{BestInModule}\underline{0.7454} & 0.6842 & \cellcolor{BestOverall}\textbf{0.7765} \\

\midrule

\multirow{2}{*}{L+J}
    & 22.89 & 12.17 & 19.16 & 21.52 & 19.98 & 15.86 & 18.66 & 18.06 & 16.14 & 17.66 & \cellcolor{BestInModule}\underline{22.72} & \cellcolor{BestOverall}\textbf{22.89} \\
& \cellcolor{BestInModule}\underline{0.8704} & 0.5433 & 0.6528 & 0.7671 & \cellcolor{BestOverall}\textbf{0.8706} & 0.6048 & 0.7205 & 0.7787 & 0.6375 & 0.8489 & 0.7635 & 0.7400 \\
\midrule

\multirow{2}{*}{L+B+N}
& 21.43 & 10.60 & 20.83 & 22.63 & 18.89 & 15.30 & \cellcolor{BestInModule}\underline{22.00} & 12.78 & 18.91 & 17.63 &  20.41 & \cellcolor{BestOverall}\textbf{24.20} \\
& 0.6612 & 0.4608 & 0.6607 & \cellcolor{BestInModule}\underline{0.6924} & 0.6705 & 0.6202 & 0.6841 & 0.5392 & 0.6499 & 0.6889 & 0.6362 & \cellcolor{BestOverall}\textbf{0.7491} \\
\midrule

\multirow{2}{*}{L+B+J}
& 19.49 & 9.57 & 19.64 & 12.40 & 18.64 & 12.46 & 13.63 & 17.39 & 16.91 & 18.64 &  \cellcolor{BestInModule}\underline{21.16} & \cellcolor{BestOverall}\textbf{25.71} \\
& 0.6582 & 0.4897 & 0.6918 & 0.5597 & 0.6751 & 0.5774 & 0.5738 & 0.7092 & 0.6761 & \cellcolor{BestInModule}\underline{0.7899} & 0.7514 & \cellcolor{BestOverall}\textbf{0.8094} \\
\midrule

\multirow{2}{*}{L+N+J}
& \cellcolor{BestInModule}\underline{26.11} & 9.64 & 21.41 & 22.71 & 20.66 & 15.34 & 16.40 & 19.13 & 18.78 & 20.09 & 26.09 & \cellcolor{BestOverall}\textbf{27.64} \\
& \cellcolor{BestOverall}\textbf{0.9496} & 0.5456 & 0.8068 & 0.8616 & \cellcolor{BestInModule}\underline{0.9338} & 0.7068 & 0.7810 & 0.9105 & 0.8122 & 0.9162 & 0.9115 & 0.9161 \\
\midrule

\multirow{2}{*}{B+N+J}
& 21.40 & 22.43 & 21.64 & 22.47 & 23.72 & 21.19 & 22.41 & 22.42 & 22.37 & \cellcolor{BestOverall}\textbf{26.72} & 24.69 & \cellcolor{BestInModule}\underline{25.46} \\
& 0.7197 & 0.6945 & 0.6577 & 0.6942 & \cellcolor{BestOverall}\textbf{0.7861} & 0.6589 & 0.6894 & 0.6980 & 0.6876 & 0.7310 & 0.7659 & \cellcolor{BestInModule}\underline{0.7787} \\ \midrule

\multirow{2}{*}{\textbf{Average}} & \cellcolor{BestInModule}\underline{24.26} & 20.27 & 21.11 & 22.49 & 23.27 & 19.15 & 21.66 & 22.18 & 22.07 & 23.57 & 23.68 & \cellcolor{BestOverall}\textbf{25.18} \\
& \cellcolor{BestInModule}\underline{0.8173} & 0.6779 & 0.6782 & 0.7209 & 0.8105 & 0.6554 & 0.7075 & 0.7426 & 0.7187 & \cellcolor{BestOverall}\textbf{0.8199} &0.7490 & 0.7729 \\

\bottomrule
\end{tabular}
\end{adjustbox}
\vspace{-14pt}
\end{table*}

\subsection{Comparative Experiment}

We first evaluate our method on the FoundIR test set, as shown in Tab.~\ref{tab:comparison}, comparing with restoration methods including Real-ESRGAN~\cite{real-esrgan}, DGUNet~\cite{dgunt}, TransWeather~\cite{transweather}, PromptIR~\cite{prompter}, DiffUIR~\cite{differ}, DA-CLIP~\cite{da-clip}, X-Restormer~\cite{x-reformer}, InstructIR~\cite{instructir}, AutoDIR~\cite{autodial}, FoundIR~\cite{foundir}. Since we develop an all-in-one image restoration model, we compare with FoundIR-Generalist (FoundIR-G), its all-in-one variant. \textbf{Trained with only 1K samples from the FoundIR training set, which accounts for merely 0.1\% to 7\% of the data used by existing approaches, our method matches or surpasses prior methods on the FoundIR test set}. Notably, compared with FoundIR models trained on 1M samples, our method even outperforms the all-in-one variant~(FoundIR-G) of FoundIR trained using 1M samples on several metrics. The qualitative results in Fig.~\ref{fig:exp:foundir-visual} further confirm its superiority. These results demonstrate remarkable data efficiency and clearly validate the effectiveness of introducing pretrained video generative priors, which substantially enhance restoration capability under severely constrained training data.

More importantly, the superiority of our method extends beyond in domain evaluation. Results on out of distribution benchmarks including Dense-Haze~\cite{dense-haze}, UHD-LL~\cite{uhd-ll}, NH-Haze~\cite{nh-haze}, UAV-Rain1K~\cite{uav-rain1k}, and HQ-NightRain~\cite{hq-nightrain}, as shown in Tab.~\ref{tab:exp:other benchmark}, demonstrate strong cross dataset generalization, where our method consistently achieves clear performance gains over competing approaches. This further confirms that video generative models implicitly capture robust and transferable visual priors, which can be effectively adapted to restoration scenarios.

We also evaluate the effectiveness of our correction module. As reported in Tab.~\ref{tab:comparison}, PSNR increases by 1.4 dB and SSIM improves by 0.024. Visual comparisons in Fig.~\ref{figure:exp:refine and ablation}(a) also demonstrate enhanced perceptual fidelity. We attribute these gains to the resolution bias of video generative models, where models are typically trained on moderate resolution data such as 720p~\cite{wan2025}, which limits the ability to recover high frequency textures. By incorporating a dedicated correction model for detail enhancement, our method decouples structural restoration from high frequency recovery, leading to improved reconstruction quality.

\begin{table*}[t]
\centering
\caption{Quantitative comparisons ($\frac{\mathrm{PSNR}}{\mathrm{SSIM}}$) on the public benchmarks. Underlining and boldface indicate the second-best and best methods, respectively.}\label{tab:exp:other benchmark}
\vspace{-0.5mm}
\resizebox{\textwidth}{!}{%

\begin{tabular}{ll|ccccccccc|c}
    
    \toprule
        \multicolumn{2}{c|}{Methods} & PromptIR & TransWeather & DA-CLIP & DiffUIR & AutoDIR & InstructIR  & X-Restormer & AgenticIR & FoundIR-G  & Ours \\ \midrule
        \multicolumn{2}{c|}{Data Scale $\downarrow$} & 77K & 19K & 53K & 30K & 23K & 80K & 40K & \cellcolor{BestInModule}\underline{15K} & 1M &  \cellcolor{BestOverall}\textbf{1K} \\ \midrule
        \multirow{2}{*}{Dense-Haze} & PSNR $\uparrow$ & 9.57 & 10.51 & 10.94 & 9.59 & \cellcolor{BestOverall}\textbf{12.33} & 11.03 & 9.57 & 10.11 & 9.29 & \cellcolor{BestInModule}\underline{11.97} \\ 
        & SSIM ~$\uparrow$ & 0.4333 & 0.4523 & 0.459 & 0.4326 & \cellcolor{BestInModule}\underline{0.4862} & 0.4649 & 0.4295 & 0.3884 & 0.4307 & \cellcolor{BestOverall}\textbf{0.4897} \\ \midrule
        \multirow{2}{*}{UHD-LL} & PSNR $\uparrow$ & 11.76 & 12.3 & 18.51 & 11.27 & \cellcolor{BestOverall}\textbf{22.52} & \cellcolor{BestInModule}\underline{20.03} & 11.56 & 12.82 & 10.61 & 17.87 \\ 
         & SSIM ~$\uparrow$ & 0.6170 & 0.6557 & \cellcolor{BestInModule}\underline{0.8120} & 0.5975 & \cellcolor{BestOverall}\textbf{0.8572} & 0.7356 & 0.6113 & 0.6649 & 0.5775 & 0.7999 \\ \midrule
        \multirow{2}{*}{NH-Haze} & PSNR $\uparrow$ & 11.38 & 11.58 & 12.35 & 11.39 & \cellcolor{BestInModule}\underline{12.71} & 12.24 & 11.36 & 12.2 & 11.43 & \cellcolor{BestOverall}\textbf{12.97} \\ 
        & SSIM ~$\uparrow$ & 0.4343 & 0.4110 & 0.4662 & 0.422 & \cellcolor{BestInModule}\underline{0.4774} & \cellcolor{BestOverall}\textbf{0.4984} & 0.4131 & 0.4495 & 0.4491 & 0.4241 \\ \midrule
        \multirow{2}{*}{UAV-Rain1k} & PSNR $\uparrow$ & 15.16 & 14.85 & 15.38 & 15.2 & \cellcolor{BestInModule}\underline{15.41} & 13.75 & 15.16 & 14.26 & 15.11 & \cellcolor{BestOverall}\textbf{15.56} \\ 
        & SSIM ~$\uparrow$ & \cellcolor{BestOverall}\textbf{0.6605} & 0.5381 & 0.6035 & 0.6389 & 0.5834 & 0.3240 & 0.6397 & 0.5300 & \cellcolor{BestInModule}\underline{0.6411} & 0.5152 \\ \midrule
        \multirow{2}{*}{HQ-NightRain} & PSNR $\uparrow$ & 10.43 & 12.88 & 14.78 & 11.58 & 11.67 & 10.92 & 10.33 & \cellcolor{BestInModule}\underline{16.29} & 11.57 & \cellcolor{BestOverall}\textbf{27.70} \\ 
        & SSIM ~$\uparrow$ & 0.4485 & 0.4774 & 0.5398 & 0.4827 & 0.4697 & 0.3972 & 0.4376 & \cellcolor{BestInModule}\underline{0.5681} & 0.5220 & \cellcolor{BestOverall}\textbf{0.8747} \\ \midrule
\end{tabular}
}
\label{tab:public}
\end{table*}

\begin{wrapfigure}[13]{r}{0.45\textwidth} 
  \vspace{-2.5em}  
  \centering
  \includegraphics[width=1.0\linewidth]{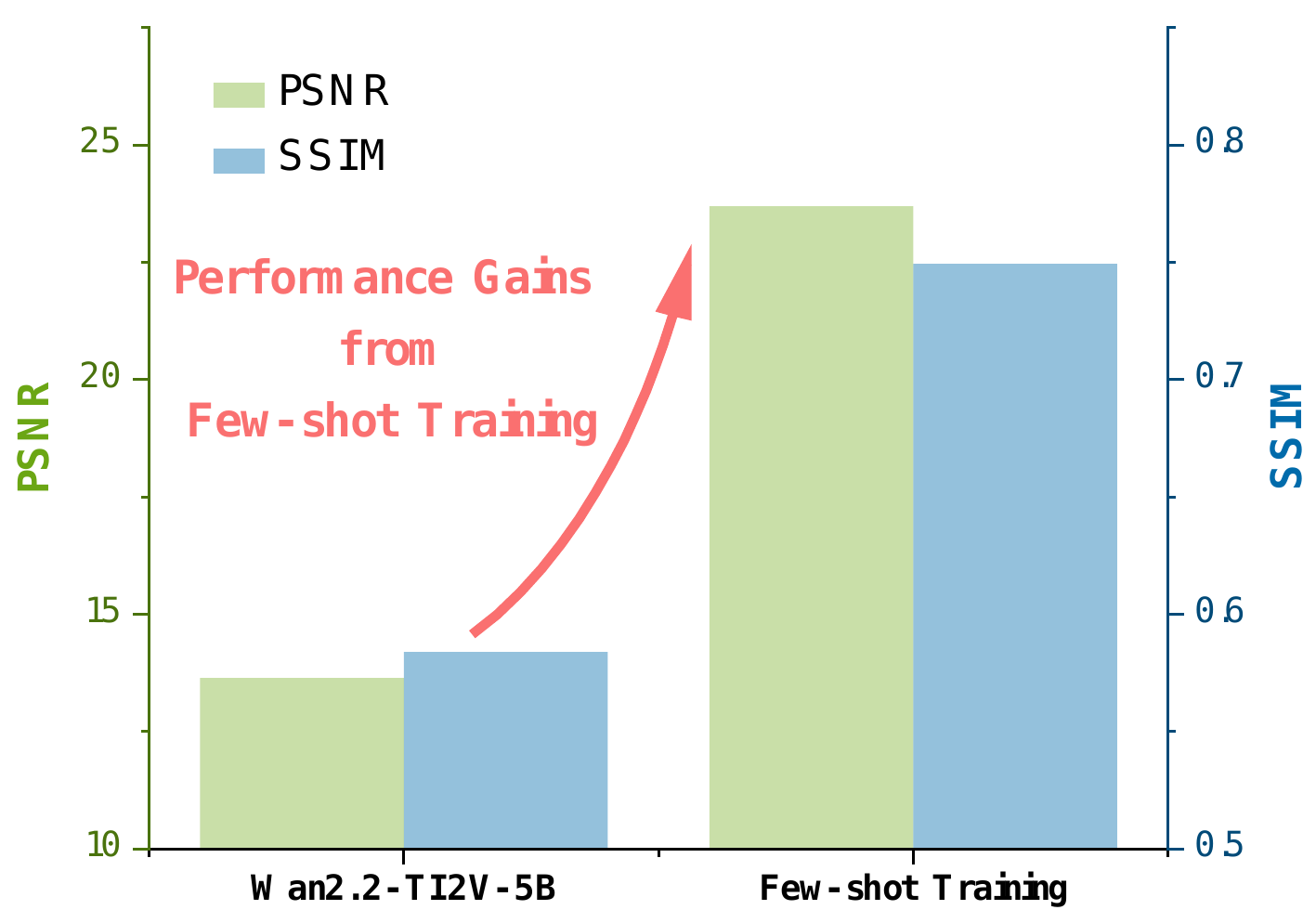}
  \vspace{-10pt}
  \caption{Performance improvement on the test dataset of FoundIR brought by few-shot training.} 
  \label{fig:training_improvement}
\end{wrapfigure}

We also show qualitative comparisons of the video generative model before and after fine tuning in Fig.\ref{fig:training_improvement}. The off the shelf model struggles to understand the restoration task, which demonstrates the necessity of our few shot training strategy. Our approach effectively unlocks the task specific potential of pretrained generative priors, enabling high fidelity restoration with strong data efficiency.

These results provide strong evidence that video generative models can serve as general visual foundation models for low-level vision tasks. By formulating restoration as a progressive generative refinement process, our approach demonstrates that high-quality restoration capability can be efficiently activated under few-shot supervision. This suggests a scalable paradigm toward universal visual prior learning, enabling effective knowledge transfer across diverse degradation types and low-level vision tasks in practice.

\subsection{Ablation Study}

\begin{figure*}[t]
\centering  
\includegraphics[width=1 \textwidth]{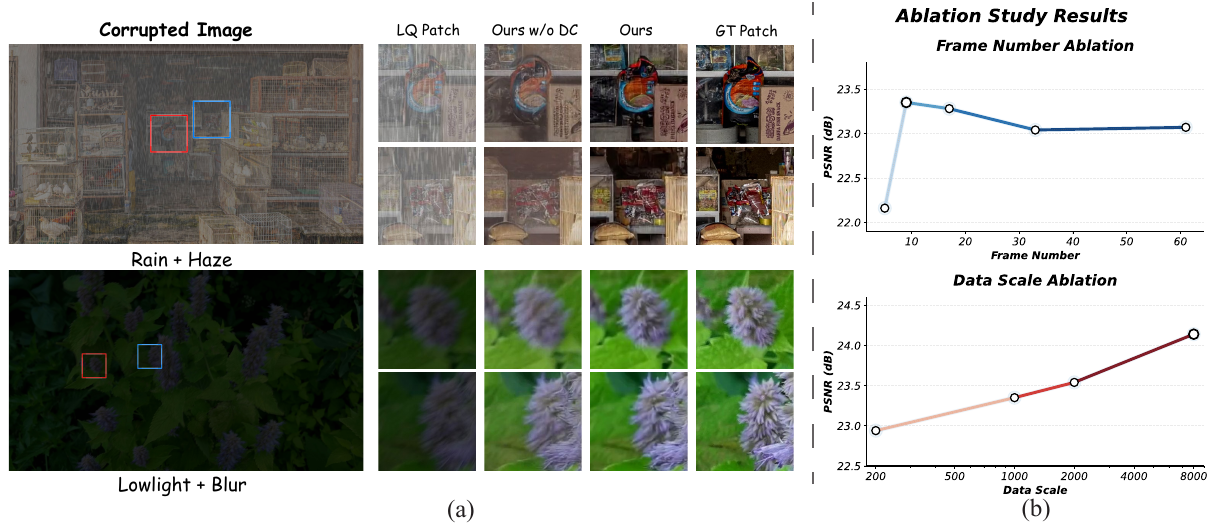} 
\vspace{-0.4cm}
\caption{(a) Comparison on the FoundIR test set with and without the refine model. GT denotes ground truth. Correction improves visual quality and enhances fine details. (b) Ablation study results. Top: effect of different training frame numbers for image restoration. Bottom: performance improves as training data scale increases. } 
\label{figure:exp:refine and ablation}  
\vspace{-2pt}
\end{figure*}

\noindent \textbf{Frame Count.}
\begin{table*}[t]
\centering
\caption{Ablation study about the number of frames ($\frac{\mathrm{PSNR}}{\mathrm{SSIM}}$, both higher is better).}
\label{tab:ablation-frame}
\vspace{-1pt}
\begin{threeparttable}
\begin{adjustbox}{width=\textwidth}
\begin{tabular}{lccccccc|ccccccccccccc|c}
\toprule
\multirow{2}{*}{\textbf{Methods}} 
& \multicolumn{7}{c|}{\textbf{Isolated Degradation}} 
& \multicolumn{13}{c|}{\textbf{Coupled Degradation}} 
& \multirow{2}{*}{\textbf{Average}} \\

\cmidrule(lr){2-8}
\cmidrule(lr){9-21}

& Blur & Noise & JPEG & Haze & Rain & Raindrop & Lowlight 
& B+N & B+J & N+J & R+H & L+H & L+R & L+B & L+N & L+J & L+B+N & L+B+J & L+N+J & B+N+J 
&  \\

\midrule

 \multirow{2}{*}{5 frames} & 21.25 & 31.20 & \cellcolor{BestInModule}\underline{26.49} & 17.20 & 25.01 & \cellcolor{BestInModule}\underline{25.29} & 19.55 & \cellcolor{BestOverall}\textbf{25.11} & \cellcolor{BestInModule}\underline{24.50} & 27.07 & 14.39 & 18.15 & 30.76 & 19.92 & 15.22 & 21.17 & 20.33 & 21.11 & 22.80 & \cellcolor{BestInModule}\underline{24.73} & 22.16  \\
 & 0.6852 & \cellcolor{BestOverall}\textbf{0.9439} & \cellcolor{BestInModule}\underline{0.8258} & 0.6119 & \cellcolor{BestInModule}\underline{0.7697} & \cellcolor{BestOverall}\textbf{0.7524} & \cellcolor{BestOverall}\textbf{0.8477} & \cellcolor{BestOverall}\textbf{0.8012} & 0.7551 & 0.8668 & 0.4491 & \cellcolor{BestInModule}\underline{0.5811} & 0.8933 & \cellcolor{BestInModule}\underline{0.6515} & 0.6256 & 0.7214 & \cellcolor{BestInModule}\underline{0.6510} & 0.7385 & 0.8971 & \cellcolor{BestInModule}\underline{0.7525} & 0.7242 \\ \midrule

 \multirow{2}{*}{9 frames} &21.42 & 33.60 & \cellcolor{BestOverall}\textbf{27.05} & 20.32 & \cellcolor{BestInModule}\underline{25.54} & \cellcolor{BestOverall}\textbf{25.62} & \cellcolor{BestOverall}\textbf{20.62} & 23.34 & 23.09 & 26.39 & 18.18 & \cellcolor{BestOverall}\textbf{19.70} & \cellcolor{BestOverall}\textbf{32.64} & \cellcolor{BestOverall}\textbf{21.24} & \cellcolor{BestOverall}\textbf{16.93} & \cellcolor{BestOverall}\textbf{22.72} & \cellcolor{BestOverall}\textbf{23.12} & \cellcolor{BestInModule}\underline{22.37} & \cellcolor{BestOverall}\textbf{25.52} & 21.54 & \cellcolor{BestOverall}\textbf{23.35} \\
 & \cellcolor{BestInModule}\underline{0.7023} & \cellcolor{BestInModule}\underline{0.9404} & \cellcolor{BestOverall}\textbf{0.8294} & \cellcolor{BestInModule}\underline{0.6833} & 0.7615 & \cellcolor{BestInModule}\underline{0.7516} & 0.8405 & 0.7550 & 0.7181 & 0.8541 & \cellcolor{BestInModule}\underline{0.5330} & \cellcolor{BestOverall}\textbf{0.5923} & \cellcolor{BestOverall}\textbf{0.9069} & \cellcolor{BestOverall}\textbf{0.6564} & 0.6879 & \cellcolor{BestOverall}\textbf{0.7594} & \cellcolor{BestOverall}\textbf{0.7115} & 0.7466 & \cellcolor{BestInModule}\underline{0.8916} & 0.6770 & \cellcolor{BestOverall}\textbf{0.7385}\\ \midrule

\multirow{2}{*}{17 frames} &  \cellcolor{BestOverall}\textbf{22.05} & \cellcolor{BestInModule}\underline{33.86} & 26.32 & \cellcolor{BestOverall}\textbf{21.32} & 24.69 & 24.51 & \cellcolor{BestInModule}\underline{20.08} & 24.41 & 24.26 & 28.36 & \cellcolor{BestInModule}\underline{18.45} & 19.23 & 32.20 & 19.78 & 16.53 & \cellcolor{BestInModule}\underline{21.62} & \cellcolor{BestInModule}\underline{20.36} & \cellcolor{BestOverall}\textbf{23.03} & \cellcolor{BestInModule}\underline{24.29} & 22.93 & \cellcolor{BestInModule}\underline{23.28} \\
& \cellcolor{BestOverall}\textbf{0.7100} & 0.9353 & 0.8166 & \cellcolor{BestOverall}\textbf{0.6849} & 0.7439 & 0.7359 & \cellcolor{BestInModule}\underline{0.8455} & 0.7790 & \cellcolor{BestInModule}\underline{0.7732} & 0.8964 & \cellcolor{BestOverall}\textbf{0.5338} & 0.5767 & 0.8976 & 0.6302 & \cellcolor{BestInModule}\underline{0.6880} & \cellcolor{BestInModule}\underline{0.7387} & 0.6337 & \cellcolor{BestOverall}\textbf{0.7851} & \cellcolor{BestOverall}\textbf{0.8977} & 0.7111 & \cellcolor{BestInModule}\underline{0.7370} \\ \midrule

 \multirow{2}{*}{33 frames} & 20.65 & \cellcolor{BestOverall}\textbf{34.04} & 26.12 & 20.60 & 24.86 & 24.52 & 19.99 & 23.14 & \cellcolor{BestOverall}\textbf{26.27} & \cellcolor{BestInModule}\underline{29.51} & 18.15 & 19.18 & \cellcolor{BestInModule}\underline{32.30} & 18.98 & \cellcolor{BestInModule}\underline{16.86} & 21.31 & 19.66 & 20.81 & 24.11 & \cellcolor{BestOverall}\textbf{25.59} & 23.04 \\
 & 0.6808 & 0.9295 & 0.8006 & 0.6665 & 0.7473 & 0.7228 & 0.8434 & 0.7483 & \cellcolor{BestOverall}\textbf{0.7937} & \cellcolor{BestInModule}\underline{0.9025} & 0.5204 & 0.5640 & \cellcolor{BestInModule}\underline{0.9034} & 0.6378 & \cellcolor{BestOverall}\textbf{0.6896} & 0.7298 & 0.6248 & \cellcolor{BestInModule}\underline{0.7571} & 0.8939 & \cellcolor{BestOverall}\textbf{0.7640} & 0.7294 \\ \midrule
 
  \multirow{2}{*}{61 frames} & \cellcolor{BestInModule}\underline{21.62} & 33.60 & 25.87 & \cellcolor{BestInModule}\underline{20.65} & \cellcolor{BestOverall}\textbf{26.25} & 24.90 & 19.33 & \cellcolor{BestInModule}\underline{24.49} & 24.12 & \cellcolor{BestOverall}\textbf{31.00} & \cellcolor{BestOverall}\textbf{19.29} & \cellcolor{BestInModule}\underline{19.42} & 31.77 & \cellcolor{BestInModule}\underline{20.25} & 12.43 & 19.70 & 20.01 & 21.63 & 23.16 & 23.04 & 23.07  \\
 & 0.6889 & 0.9360 & 0.7971 & 0.6457 & \cellcolor{BestOverall}\textbf{0.7842} & 0.7253 & 0.8024 & \cellcolor{BestInModule}\underline{0.7807} & 0.7489 & \cellcolor{BestOverall}\textbf{0.9164} & 0.5300 & 0.5721 & 0.8996 & 0.6091 & 0.5589 & 0.7064 & 0.6181 & 0.7326 & 0.8800 & 0.7213 & 0.7199 \\

\bottomrule
\end{tabular}
\end{adjustbox}
\end{threeparttable}
\end{table*}
We analyze the impact of frame numbers in the video generation process for image restoration. As shown in Tab.\ref{tab:ablation-frame} and Fig.\ref{figure:exp:refine and ablation}(b), we evaluate frame numbers ranging from 5 to 61 on the FoundIR test set. Interestingly, performance does not monotonically improve with more frames. Instead, the 9-frame setting consistently yields better results than 33 and 61 frames.

This observation suggests that the generative prior learned by the video model primarily emphasizes global structural coherence rather than fine grained temporal modeling. In this view, image restoration can be formulated as a conditional generative process guided by strong semantic and geometric priors, where missing content is inferred through high level structural reasoning. Therefore, increasing the number of frames does not necessarily provide additional informative signals, as neighboring frames tend to become semantically redundant under smooth restoration trajectories in the latent space. Consequently, a moderate number of frames is sufficient for approximating the restoration process, while excessive temporal sampling may introduce redundant constraints and increase computational burden without significant performance gains. This further indicates that restoration performance is mainly determined by learned spatial semantic priors rather than dense temporal sampling.

\noindent \textbf{Progressive Curriculum Training.}
\begin{table*}[t]
\centering
\caption{Ablation study on Progressive Curriculum Training ($\frac{\mathrm{PSNR}}{\mathrm{SSIM}}$).}
\label{tab:ablation-training-process}
\vspace{-0pt}
\begin{threeparttable}
\begin{adjustbox}{width=\textwidth}
\begin{tabular}{l|ccccccc|ccccccccccccc|c}
\toprule
\multirow{2}{*}{\textbf{Resolution}} 
& \multicolumn{7}{c|}{\textbf{Isolated Degradation}} 
& \multicolumn{13}{c|}{\textbf{Coupled Degradation}} 
& \multirow{2}{*}{\textbf{Average}} \\

\cmidrule(lr){2-8}
\cmidrule(lr){9-21}

& Blur & Noise & JPEG & Haze & Rain & Raindrop & Lowlight 
& B+N & B+J & N+J & R+H & L+H & L+R & L+B & L+N & L+J & L+B+N & L+B+J & L+N+J & B+N+J 
&  \\

\midrule

 \multirow{2}{*}{512} & 21.25 & 31.20 & 26.49 & 17.20 & 25.01 & 25.29 & 19.55 & \cellcolor{BestOverall}\textbf{25.11} & 24.50 & 27.07 & 14.39 & 18.15 & 30.76 & 19.92 & 15.22 & 21.17 & 20.33 & 21.11 & 22.80 & \cellcolor{BestInModule}\underline{24.73} & 22.16  \\
 & 0.6852 & 0.9439 & 0.8258 & 0.6119 & 0.7697 & 0.7524 & \cellcolor{BestInModule}\underline{0.8477} & \cellcolor{BestOverall}\textbf{0.8012} & 0.7551 & 0.8668 & 0.4491 & 0.5811 & 0.8933 & \cellcolor{BestInModule}\underline{0.6515} & 0.6256 & 0.7214 & 0.6510 & 0.7385 & 0.8971 & 0.7525 & 0.7242   \\ \midrule

 \multirow{2}{*}{720} &  \cellcolor{BestOverall}\textbf{22.33} & \cellcolor{BestInModule}\underline{33.64} & 27.97 & 20.64 & 25.76 & 25.63 & 19.34 & 22.24 & 23.21 & 28.48 & \cellcolor{BestOverall}\textbf{19.79} & \cellcolor{BestInModule}\underline{19.57} & 32.06 & 19.18 & \cellcolor{BestInModule}\underline{17.36} & 22.63 & \cellcolor{BestInModule}\underline{22.10} & 21.95 & 24.83 & 21.45 & 23.35 \\
 &  \cellcolor{BestOverall}\textbf{0.7253} & \cellcolor{BestInModule}\underline{0.9440} & 0.8464 & 0.6854 & 0.7792 & 0.7623 & 0.8363 & 0.7275 & 0.7131 & 0.8869 & \cellcolor{BestOverall}\textbf{0.5594} & \cellcolor{BestInModule}\underline{0.5948} & 0.9010 & 0.6411 & \cellcolor{BestOverall}\textbf{0.7036} & 0.7631 & \cellcolor{BestOverall}\textbf{0.7030} & 0.7368 & 0.9055 & 0.6823 & 0.7442  \\ \midrule

\multirow{2}{*}{960} &   21.55 & \cellcolor{BestOverall}\textbf{34.40} & 28.10 & 20.55 & \cellcolor{BestInModule}\underline{26.37} & 25.47 & 19.45 & 23.42 & \cellcolor{BestOverall}\textbf{25.87} & \cellcolor{BestOverall}\textbf{33.60} & \cellcolor{BestInModule}\underline{18.73} & 18.01 & 32.18 & 18.81 & 17.18 & 22.25 & 19.50 & 20.48 & \cellcolor{BestInModule}\underline{25.82} & 22.85 & 23.41 \\
 & \cellcolor{BestInModule}\underline{0.7089} & \cellcolor{BestOverall}\textbf{0.9454} & \cellcolor{BestInModule}\underline{0.8477} & \cellcolor{BestInModule}\underline{0.6867} & \cellcolor{BestOverall}\textbf{0.7970} & \cellcolor{BestOverall}\textbf{0.7700} & 0.8294 & 0.7657 & \cellcolor{BestOverall}\textbf{0.7937} & \cellcolor{BestOverall}\textbf{0.9440} & 0.5570 & 0.5808 & \cellcolor{BestInModule}\underline{0.9112} & 0.6404 & 0.6782 & 0.7555 & 0.6312 & 0.7402 & 0.9097 & 0.7167 & 0.7474 \\ \midrule

 \multirow{2}{*}{512+720} & \cellcolor{BestInModule}\underline{21.70} & 30.71 & \cellcolor{BestInModule}\underline{28.18} & \cellcolor{BestInModule}\underline{21.08} & 25.76 & \cellcolor{BestOverall}\textbf{25.93} & \cellcolor{BestInModule}\underline{19.89} & \cellcolor{BestInModule}\underline{24.29} & 25.55 & 28.99 & 18.61 & 19.13 & \cellcolor{BestInModule}\underline{32.71} & \cellcolor{BestOverall}\textbf{20.83} & \cellcolor{BestOverall}\textbf{17.62} & \cellcolor{BestInModule}\underline{22.65} & \cellcolor{BestOverall}\textbf{22.45} & \cellcolor{BestOverall}\textbf{23.86} & 25.59 & 23.11 & \cellcolor{BestInModule}\underline{23.59}  \\
 &  0.6999 & 0.9394 & 0.8469 & 0.6842 & 0.7693 & 0.7640 & 0.8256 & 0.7718 & 0.7900 & 0.8876 & 0.5377 & 0.5862 & 0.9103 & 0.6438 & 0.6840 & \cellcolor{BestOverall}\textbf{0.7687} & \cellcolor{BestInModule}\underline{0.6852} & \cellcolor{BestOverall}\textbf{0.7832} & \cellcolor{BestInModule}\underline{0.9101} & 0.7117 & \cellcolor{BestInModule}\underline{0.7451} \\ \midrule
 
  \multirow{2}{*}{720+512} &21.38 & 32.32 & 26.87 & 19.43 & 25.36 & 25.40 & \cellcolor{BestOverall}\textbf{19.95} & 24.15 & \cellcolor{BestInModule}\underline{25.78} & 26.40 & 16.01 & 18.74 & 31.96 & 20.03 & 17.23 & 22.03 & 20.26 & \cellcolor{BestInModule}\underline{22.16} & 23.85 & \cellcolor{BestOverall}\textbf{24.77} & 22.90\\
 &  0.6951 & 0.9416 & 0.8341 & 0.6612 & 0.7683 & 0.7596 & \cellcolor{BestOverall}\textbf{0.8394} & \cellcolor{BestInModule}\underline{0.7730} & \cellcolor{BestInModule}\underline{0.7790} & 0.8572 & 0.4952 & 0.5742 & 0.9024 & 0.6430 & 0.6915 & 0.7512 & 0.6198 & \cellcolor{BestInModule}\underline{0.7637} & 0.9011 & \cellcolor{BestInModule}\underline{0.7541} & 0.7355 \\ \midrule
 
   \multirow{2}{*}{512+720+960} &21.02 & 32.61 & \cellcolor{BestOverall}\textbf{28.21} & \cellcolor{BestOverall}\textbf{21.25} & \cellcolor{BestOverall}\textbf{26.40} & \cellcolor{BestInModule}\underline{25.77} & 19.18 & 23.80 & 24.05 & \cellcolor{BestInModule}\underline{33.33} & 18.70 & \cellcolor{BestOverall}\textbf{19.74} & \cellcolor{BestOverall}\textbf{32.90} & \cellcolor{BestInModule}\underline{20.37} & \cellcolor{BestInModule}\underline{17.36} & \cellcolor{BestOverall}\textbf{22.72} & 20.41 & 21.16 & \cellcolor{BestOverall}\textbf{26.09} & 24.69 & \cellcolor{BestOverall}\textbf{23.68} \\
 & 0.6862 & 0.9397 & \cellcolor{BestOverall}\textbf{0.8487} & \cellcolor{BestOverall}\textbf{0.6918} & \cellcolor{BestInModule}\underline{0.7957} & \cellcolor{BestInModule}\underline{0.7698} & 0.8278 & 0.7703 & 0.7571 & \cellcolor{BestInModule}\underline{0.9422} & \cellcolor{BestInModule}\underline{0.5574} & \cellcolor{BestOverall}\textbf{0.5983} & \cellcolor{BestOverall}\textbf{0.9126} & \cellcolor{BestOverall}\textbf{0.6543} & 0.6842 & \cellcolor{BestInModule}\underline{0.7635} & 0.6362 & 0.7514 & \cellcolor{BestOverall}\textbf{0.9115} & \cellcolor{BestOverall}\textbf{0.7659} & \cellcolor{BestOverall}\textbf{0.7490}\\

\bottomrule
\end{tabular}
\end{adjustbox}
\end{threeparttable}
\end{table*}
We study the effectiveness of the proposed progressive curriculum training strategy, as shown in Tab.~\ref{tab:ablation-training-process}, where we compare fixed resolution training with different progressive schedules. Here, numbers denote the training resolution (e.g., 512 indicates resizing the input image such that the shorter side is aligned to 512 while preserving the aspect ratio, followed by 512×512 random cropping). All models are trained for 300 epochs in total, with epochs evenly distributed across stages for multi-stage training.

The results show that performance consistently improves with increasing resolution, since higher resolutions provide richer high frequency structural and textural cues for learning more discriminative restoration priors. Moreover, progressive resolution training outperforms single resolution training (e.g., 512+720 and 512+720+960), suggesting that a coarse to fine optimization paradigm better matches the hierarchical nature of image restoration. In contrast, decreasing resolution schedules lead to performance degradation~(e.g. 720+512), as the model tends to prioritize global appearance consistency while sacrificing fine detail modeling. Overall, the proposed strategy consistently enhances both training stability and restoration quality by better aligning optimization with the intrinsic coarse to fine restoration process.

\subsection{Discussions}

\noindent \textbf{Few-shot Capability.}\label{sec:exp:analysis:few-shot}
We discuss the effectiveness of leveraging video generation priors for few shot image restoration. As shown in Tab.~\ref{tab:comparison}, our method achieves competitive or even superior performance while using significantly less training data. Specifically, our model is trained using only 0.1\% – 7\% of the data required by conventional methods, demonstrating that large scale video generative pretraining captures strong generic structural and motion aware priors that can be transferred to restoration tasks. From a learning perspective, our approach focuses on activating and adapting pretrained generative knowledge with limited task specific image restoration supervision, reducing data dependency while preserving model expressiveness.

\begin{table*}[t]
\centering
\caption{Ablation study about the scale of training data ($\frac{\mathrm{PSNR}}{\mathrm{SSIM}}$, both higher is better).}
\label{tab:training-scale}
\vspace{-3pt}
\begin{threeparttable}
\begin{adjustbox}{width=\textwidth}
\begin{tabular}{lccccccc|ccccccccccccc|c}
\toprule
\multirow{2}{*}{\textbf{Data Scale}} 
& \multicolumn{7}{c|}{\textbf{Isolated Degradation}} 
& \multicolumn{13}{c|}{\textbf{Coupled Degradation}} 
& \multirow{2}{*}{\textbf{Average}} \\

\cmidrule(lr){2-8}
\cmidrule(lr){9-21}

& Blur & Noise & JPEG & Haze & Rain & Raindrop & Lowlight 
& B+N & B+J & N+J & R+H & L+H & L+R & L+B & L+N & L+J & L+B+N & L+B+J & L+N+J & B+N+J 
&  \\

\midrule

 \multirow{2}{*}{0.2K} &22.14 & \cellcolor{BestOverall}\textbf{33.90} & 27.01 & 19.64 & 25.96 & 24.34 & 20.21 & 22.49 & 22.03 & \cellcolor{BestInModule}\underline{30.06} & 18.64 & 18.52 & 30.37 & 21.39 & \cellcolor{BestInModule}\underline{16.85} & 19.41 & 21.85 & 20.37 & 23.91 & 21.85 & 22.94 \\
 &0.7213 & \cellcolor{BestOverall}\textbf{0.9463} & 0.8340 & 0.6568 & 0.7964 & 0.7540 & 0.8548 & 0.7380 & 0.6789 & \cellcolor{BestOverall}\textbf{0.9156} & 0.5297 & 0.5822 & 0.8931 & 0.6811 & 0.6414 & 0.7064 & \cellcolor{BestInModule}\underline{0.7109} & 0.6582 & 0.9116 & \cellcolor{BestInModule}\underline{0.6954} & 0.7360 \\ \midrule

 \multirow{2}{*}{1K} &   \cellcolor{BestInModule}\underline{22.33} & \cellcolor{BestInModule}\underline{33.64} & \cellcolor{BestOverall}\textbf{27.97} & \cellcolor{BestInModule}\underline{20.64} & 25.76 & \cellcolor{BestInModule}\underline{25.63} & 19.34 & 22.24 & 23.21 & 28.48 & \cellcolor{BestInModule}\underline{19.79} & 19.57 & \cellcolor{BestInModule}\underline{32.06} & 19.18 & \cellcolor{BestOverall}\textbf{17.36} & \cellcolor{BestOverall}\textbf{22.63} & \cellcolor{BestInModule}\underline{22.10} & \cellcolor{BestOverall}\textbf{21.95} & 24.83 & 21.45 & 23.35 \\
 &  \cellcolor{BestInModule}\underline{0.7253} & 0.9440 & \cellcolor{BestOverall}\textbf{0.8464} & \cellcolor{BestOverall}\textbf{0.6854} & 0.7792 & \cellcolor{BestOverall}\textbf{0.7623} & 0.8363 & 0.7275 & 0.7131 & 0.8869 & \cellcolor{BestOverall}\textbf{0.5594} & 0.5948 & 0.9010 & 0.6411 & \cellcolor{BestOverall}\textbf{0.7036} & \cellcolor{BestOverall}\textbf{0.7631} & 0.7030 & 0.7368 & 0.9055 & 0.6823 & 0.7442  \\ \midrule

\multirow{2}{*}{2K} &  21.92 & 31.98 & 27.50 & 19.64 & \cellcolor{BestOverall}\textbf{26.73} & \cellcolor{BestOverall}\textbf{25.74} & \cellcolor{BestInModule}\underline{20.63} & \cellcolor{BestOverall}\textbf{25.79} & \cellcolor{BestInModule}\underline{25.18} & 28.86 & 18.37 & \cellcolor{BestInModule}\underline{19.81} & \cellcolor{BestOverall}\textbf{32.25} & \cellcolor{BestInModule}\underline{21.65} & 16.08 & \cellcolor{BestInModule}\underline{22.26} & 22.08 & 21.01 & \cellcolor{BestInModule}\underline{26.28} & \cellcolor{BestInModule}\underline{23.79} & \cellcolor{BestInModule}\underline{23.54} \\
 & 0.7082 & 0.9432 & 0.8361 & 0.6501 & \cellcolor{BestOverall}\textbf{0.8003} & \cellcolor{BestInModule}\underline{0.7611} & \cellcolor{BestOverall}\textbf{0.8630} & \cellcolor{BestOverall}\textbf{0.8199} & \cellcolor{BestInModule}\underline{0.7718} & 0.8930 & 0.5273 & \cellcolor{BestInModule}\underline{0.6120} & \cellcolor{BestOverall}\textbf{0.9081} & \cellcolor{BestInModule}\underline{0.6822} & \cellcolor{BestInModule}\underline{0.6486} & \cellcolor{BestInModule}\underline{0.7367} & 0.6946 & \cellcolor{BestInModule}\underline{0.7447} & \cellcolor{BestInModule}\underline{0.9135} & \cellcolor{BestOverall}\textbf{0.7279} & \cellcolor{BestInModule}\underline{0.7469} \\ \midrule

 \multirow{2}{*}{8K} &\cellcolor{BestOverall}\textbf{22.98} & 33.41 & \cellcolor{BestInModule}\underline{27.57} & \cellcolor{BestOverall}\textbf{21.18} & \cellcolor{BestInModule}\underline{26.54} & 25.11 & \cellcolor{BestOverall}\textbf{21.49} & \cellcolor{BestInModule}\underline{25.10} & \cellcolor{BestOverall}\textbf{26.02} & \cellcolor{BestOverall}\textbf{30.64} & \cellcolor{BestOverall}\textbf{19.97} & \cellcolor{BestOverall}\textbf{20.00} & 31.67 & \cellcolor{BestOverall}\textbf{21.76} & 16.28 & 21.14 & \cellcolor{BestOverall}\textbf{22.86} & \cellcolor{BestInModule}\underline{21.86} & \cellcolor{BestOverall}\textbf{28.27} & \cellcolor{BestOverall}\textbf{23.91} & \cellcolor{BestOverall}\textbf{24.14} \\
 & \cellcolor{BestOverall}\textbf{0.7284} & \cellcolor{BestInModule}\underline{0.9454} & \cellcolor{BestInModule}\underline{0.8375} & \cellcolor{BestInModule}\underline{0.6704} & \cellcolor{BestInModule}\underline{0.7982} & \cellcolor{BestInModule}\underline{0.7611} & \cellcolor{BestInModule}\underline{0.8628} & \cellcolor{BestInModule}\underline{0.7999} & \cellcolor{BestOverall}\textbf{0.7827} & \cellcolor{BestInModule}\underline{0.9088} & \cellcolor{BestInModule}\underline{0.5562} & \cellcolor{BestOverall}\textbf{0.6200} & \cellcolor{BestInModule}\underline{0.9063} & \cellcolor{BestOverall}\textbf{0.7153} & 0.6458 & 0.7214 & \cellcolor{BestOverall}\textbf{0.7354} & \cellcolor{BestOverall}\textbf{0.7505} & \cellcolor{BestOverall}\textbf{0.9137} & 0.6786 & \cellcolor{BestOverall}\textbf{0.7549} \\

\bottomrule
\end{tabular}
\end{adjustbox}
\end{threeparttable}
\end{table*}




\begin{table*}[t]
\centering
\caption{Quantitative comparisons ($\frac{\mathrm{PSNR}}{\mathrm{SSIM}}$) on the unseen task for our method about desnowing. Underlining and boldface indicate the second-best and best methods, respectively. FoundIR-G denotes the generalist version of FoundIR.}\label{tab:exp:unseen}

\resizebox{\textwidth}{!}{%

\begin{tabular}{ll|ccccccc|c}
    
    \toprule
        \multicolumn{2}{c|}{Methods} & PromptIR & TransWeather & DA-CLIP & DiffUIR & X-Restormer & AgenticIR & FoundIR-G  & Ours \\ \midrule
        \multicolumn{2}{c|}{Data Scale $\downarrow$} & 77K & 19K & 53K & 30K & 40K & \cellcolor{BestInModule}\underline{15K} & 1M &  \cellcolor{BestOverall}\textbf{1K} \\ \midrule
        \multirow{2}{*}{WeatherBench} & PSNR $\uparrow$ & 21.54 & 20.94 & \cellcolor{BestInModule}\underline{21.59} & \cellcolor{BestOverall}\textbf{21.68} & 21.57 & 20.35 & 21.57 & 20.88 \\ 
        & SSIM ~$\uparrow$ & 0.7767 & 0.7579 & 0.7731 & \cellcolor{BestInModule}\underline{0.7795} & \cellcolor{BestOverall}\textbf{0.7813} & 0.7304 &0.7794 & 0.7105 \\ \midrule
\end{tabular}
}
\vspace{-5mm}
\end{table*}

As shown in Tab.\ref{tab:training-scale} and Fig.\ref{figure:exp:refine and ablation}, the performance of our method consistently improves as the amount of training data increases. For a fair comparison, all models are trained for 300 epochs at a fixed 720 resolution. Notably, using only 200 training samples, our method already achieves performance comparable to existing full-data baselines, highlighting strong data efficiency and generalization capability. This suggests that the pretrained video generative prior acts as a powerful regularization mechanism, enabling robust restoration under limited supervision while allowing further performance gains as more data becomes available. However, it is worth noting that expanding the training set does not consistently improve performance across all degradation types, suggesting a potential trade-off or balance during model learning.

Overall, these results further validate the strong potential of exploiting video generative priors for few-shot image restoration, suggesting that large-scale generative pretraining can serve as a general knowledge foundation for data-efficient visual learning. By transferring high-level semantic, structural, and motion-aware priors from video, generative models provide a promising paradigm for building scalable and data-efficient visual intelligence that generalizes beyond restoration tasks to broader visual understanding and synthesis problems.

\noindent \textbf{Generalization Capability.}\label{sec:exp:analysis:ood} As shown in Tab.~\ref{tab:exp:other benchmark}, although trained on only a 1K subset of FoundIR training data, our model generalizes well to multiple external benchmarks and achieves performance comparable to top-tier methods, demonstrating strong data efficiency and the feasibility of competitive restoration without large scale task specific supervision.

\begin{figure*}[t]
\centering  
\includegraphics[width=1 \textwidth]{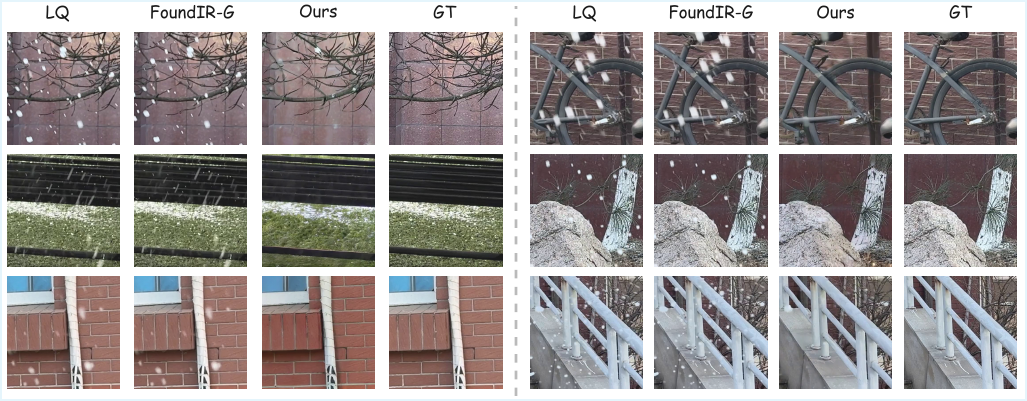} 
\vspace{-0.3cm}
\caption{Visualization results on an OOD snow removal scenario. FoundIR-G denotes the Generalist model of FoundIR. GT denotes ground truth. Both methods are evaluated under OOD conditions. Although our model is trained with only 1K samples, it shows significantly better generalization than FoundIR-G trained with 1M images.} 
\label{fig:snow}  
\vspace{-6pt}
\end{figure*}

Furthermore, Fig.~\ref{fig:snow} and Tab.~\ref{tab:exp:unseen} present qualitative and quantitative results on snow removal~\cite{weatherbench}, an unseen task for our model. Although trained with only limited image restoration data and without additional training on new degradation types, the model generalizes effectively to unseen corruptions and successfully removes snow artifacts while producing visually coherent and competitive results. This demonstrates that the model can learn transferable restoration priors and directly transfer knowledge learned during pretraining to new tasks, such as applying its understanding of snow patterns to desnowing tasks. This capability reflects the strong inductive power of video generation models, which can implicitly capture rich semantic and structural knowledge during pretraining. It is particularly important for few-shot image restoration, where minimal training data is sufficient to enable universal degradation modeling. Overall, these results highlight the strong generalization potential of generative foundation models, showing that minimal supervision can unlock broad cross-domain adaptability and pave the way toward more general visual intelligence.

\section{Conclusion}
\label{sec:con}
In this paper, we introduced V-Bridge, a framework that unlocks the restoration capability of pretrained video generative models with few-shot training. By reformulating restoration as a progressive generative refinement process and exploiting chain-like frame reasoning, we show that video foundation models can serve as powerful and transferable visual priors. Remarkably, our approach activates strong restoration capability using only a very small number of task-specific samples, further demonstrating the broad potential of video generative models beyond video synthesis. We also propose efficient progressive resolution training and lightweight refinement strategies to achieve high-fidelity restoration under limited supervision. This work provides new evidence that video generative models can serve as general visual foundation models and opens new directions for unified generative modeling in vision tasks.

\clearpage  


%
%
\bibliographystyle{splncs04}
\bibliography{main}

@String(CVPR  = {IEEE Conf. Comput. Vis. Pattern Recog.})

@String(ECCV  = {Eur. Conf. Comput. Vis.})

@String(AAAI  = {AAAI})

@String(ICIP  = {IEEE Int. Conf. Image Process.})

@String(CVPR  = {CVPR})

@String(ECCV  = {ECCV})

@String(ICIP  = {ICIP})

@inproceedings{foundir,
  title={Foundir: Unleashing million-scale training data to advance foundation models for image restoration},
  author={Li, Hao and Chen, Xiang and Dong, Jiangxin and Tang, Jinhui and Pan, Jinshan},
  booktitle={Proceedings of the IEEE/CVF international conference on computer vision},
  pages={12626--12636},
  year={2025}
}

@inproceedings{real-ce,
  title={A benchmark for chinese-english scene text image super-resolution},
  author={Ma, Jianqi and Liang, Zhetong and Xiang, Wangmeng and Yang, Xi and Zhang, Lei},
  booktitle={Proceedings of the IEEE/CVF international conference on computer vision},
  pages={19452--19461},
  year={2023}
}

@article{wan2025,
      title={Wan: Open and Advanced Large-Scale Video Generative Models}, 
      author={Team Wan and Ang Wang and Baole Ai and Bin Wen and Chaojie Mao and Chen-Wei Xie and Di Chen and Feiwu Yu and Haiming Zhao and Jianxiao Yang and Jianyuan Zeng and Jiayu Wang and Jingfeng Zhang and Jingren Zhou and Jinkai Wang and Jixuan Chen and Kai Zhu and Kang Zhao and Keyu Yan and Lianghua Huang and Mengyang Feng and Ningyi Zhang and Pandeng Li and Pingyu Wu and Ruihang Chu and Ruili Feng and Shiwei Zhang and Siyang Sun and Tao Fang and Tianxing Wang and Tianyi Gui and Tingyu Weng and Tong Shen and Wei Lin and Wei Wang and Wei Wang and Wenmeng Zhou and Wente Wang and Wenting Shen and Wenyuan Yu and Xianzhong Shi and Xiaoming Huang and Xin Xu and Yan Kou and Yangyu Lv and Yifei Li and others},
      journal = {arXiv preprint arXiv:2503.20314},
      year={2025}
}

@inproceedings{dense-haze,
  title={Dense-haze: A benchmark for image dehazing with dense-haze and haze-free images},
  author={Ancuti, Codruta O and Ancuti, Cosmin and Sbert, Mateu and Timofte, Radu},
  booktitle={2019 IEEE international conference on image processing (ICIP)},
  pages={1014--1018},
  year={2019},
  organization={IEEE}
}

@inproceedings{uhd-ll,
  title={Ultra-high-definition low-light image enhancement: A benchmark and transformer-based method},
  author={Wang, Tao and Zhang, Kaihao and Shen, Tianrun and Luo, Wenhan and Stenger, Bjorn and Lu, Tong},
  booktitle={Proceedings of the AAAI conference on artificial intelligence},
  volume={37},
  number={3},
  pages={2654--2662},
  year={2023}
}

@inproceedings{nh-haze,
author = {Codruta O. Ancuti and Cosmin Ancuti and Radu Timofte},
title = {{NH-HAZE:} An Image Dehazing Benchmark with Non-Homogeneous Hazy and Haze-Free Images},
booktitle ={Proceedings of the IEEE Conference on Computer Vision and Pattern Recognition Workshops},
series = {IEEE CVPR 2020},
year = {2020},
location = {Washington, US},
}

@inproceedings{uav-rain1k,
  title={Uav-rain1k: A benchmark for raindrop removal from uav aerial imagery},
  author={Chang, Wenhui and Chen, Hongming and He, Xin and Chen, Xiang and Shen, Liangduo},
  booktitle={Proceedings of the IEEE/CVF Conference on Computer Vision and Pattern Recognition},
  pages={15--22},
  year={2024}
}

@article{hq-nightrain,
  title={Towards unified deep image deraining: A survey and a new benchmark},
  author={Chen, Xiang and Pan, Jinshan and Dong, Jiangxin and Tang, Jinhui},
  journal={IEEE Transactions on Pattern Analysis and Machine Intelligence},
  year={2025},
  publisher={IEEE}
}

@inproceedings{weatherbench,
  title={Weatherbench: A real-world benchmark dataset for all-in-one adverse weather image restoration},
  author={Guan, Qiyuan and Yang, Qianfeng and Chen, Xiang and Song, Tianyu and Jin, Guiyue and Jin, Jiyu},
  booktitle={Proceedings of the 33rd ACM international conference on multimedia},
  pages={12607--12613},
  year={2025}
}

@inproceedings{real-esrgan,
  title={Real-esrgan: Training real-world blind super-resolution with pure synthetic data},
  author={Wang, Xintao and Xie, Liangbin and Dong, Chao and Shan, Ying},
  booktitle={Proceedings of the IEEE/CVF international conference on computer vision},
  pages={1905--1914},
  year={2021}
}

@inproceedings{dgunt,
  title={Deep generalized unfolding networks for image restoration},
  author={Mou, Chong and Wang, Qian and Zhang, Jian},
  booktitle={Proceedings of the IEEE/CVF conference on computer vision and pattern recognition},
  pages={17399--17410},
  year={2022}
}

@inproceedings{transweather,
  title={Transweather: Transformer-based restoration of images degraded by adverse weather conditions},
  author={Valanarasu, Jeya Maria Jose and Yasarla, Rajeev and Patel, Vishal M},
  booktitle={Proceedings of the IEEE/CVF conference on computer vision and pattern recognition},
  pages={2353--2363},
  year={2022}
}

@article{prompter,
  title={Promptir: Prompting for all-in-one image restoration},
  author={Potlapalli, Vaishnav and Zamir, Syed Waqas and Khan, Salman H and Shahbaz Khan, Fahad},
  journal={Advances in neural information processing systems},
  volume={36},
  pages={71275--71293},
  year={2023}
}

@inproceedings{differ,
  title={Selective hourglass mapping for universal image restoration based on diffusion model},
  author={Zheng, Dian and Wu, Xiao-Ming and Yang, Shuzhou and Zhang, Jian and Hu, Jian-Fang and Zheng, Wei-Shi},
  booktitle={Proceedings of the IEEE/CVF conference on computer vision and pattern recognition},
  pages={25445--25455},
  year={2024}
}

@inproceedings{da-clip,
title={Controlling Vision-Language Models for Multi-Task Image Restoration},
author={Ziwei Luo and Fredrik K. Gustafsson and Zheng Zhao and Jens Sj{\"o}lund and Thomas B. Sch{\"o}n},
booktitle={The Twelfth International Conference on Learning Representations},
year={2024},
url={https://openreview.net/forum?id=t3vnnLeajU}
}

@inproceedings{x-reformer,
  title={A comparative study of image restoration networks for general backbone network design},
  author={Chen, Xiangyu and Li, Zheyuan and Pu, Yuandong and Liu, Yihao and Zhou, Jiantao and Qiao, Yu and Dong, Chao},
  booktitle={European Conference on Computer Vision},
  pages={74--91},
  year={2024},
  organization={Springer}
}

@inproceedings{instructir,
  title={InstructIR: High-Quality Image Restoration Following Human Instructions},
  author={Conde, Marcos V and Geigle, Gregor and Timofte, Radu},
  booktitle    = {Proceedings of the European Conference on Computer Vision (ECCV)},
  year={2024}
}

@inproceedings{autodial,
  title={Autodir: Automatic all-in-one image restoration with latent diffusion},
  author={Jiang, Yitong and Zhang, Zhaoyang and Xue, Tianfan and Gu, Jinwei},
  booktitle={European Conference on Computer Vision},
  pages={340--359},
  year={2024},
  organization={Springer}
}

@article{cof,
  title={Video models are zero-shot learners and reasoners},
  author={Wiedemer, Thadd{\"a}us and Li, Yuxuan and Vicol, Paul and Gu, Shixiang Shane and Matarese, Nick and Swersky, Kevin and Kim, Been and Jaini, Priyank and Geirhos, Robert},
  journal={arXiv preprint arXiv:2509.20328},
  year={2025}
}

@article{ir,
  title={Digital image restoration},
  author={Banham, Mark R and Katsaggelos, Aggelos K},
  journal={IEEE signal processing magazine},
  volume={14},
  number={2},
  pages={24--41},
  year={2002},
  publisher={IEEE}
}

@inproceedings{swinir,
  title={Swinir: Image restoration using swin transformer},
  author={Liang, Jingyun and Cao, Jiezhang and Sun, Guolei and Zhang, Kai and Van Gool, Luc and Timofte, Radu},
  booktitle={Proceedings of the IEEE/CVF international conference on computer vision},
  pages={1833--1844},
  year={2021}
}

@article{seeddance,
  title={Seedance 1.0: Exploring the boundaries of video generation models},
  author={Gao, Yu and Guo, Haoyuan and Hoang, Tuyen and Huang, Weilin and Jiang, Lu and Kong, Fangyuan and Li, Huixia and Li, Jiashi and Li, Liang and Li, Xiaojie and others},
  journal={arXiv preprint arXiv:2506.09113},
  year={2025}
}

@article{opensora,
  title={Open-sora: Democratizing efficient video production for all},
  author={Zheng, Zangwei and Peng, Xiangyu and Yang, Tianji and Shen, Chenhui and Li, Shenggui and Liu, Hongxin and Zhou, Yukun and Li, Tianyi and You, Yang},
  journal={arXiv preprint arXiv:2412.20404},
  year={2024}
}

@article{hunyuanvideo,
  title={Hunyuanvideo: A systematic framework for large video generative models},
  author={Kong, Weijie and Tian, Qi and Zhang, Zijian and Min, Rox and Dai, Zuozhuo and Zhou, Jin and Xiong, Jiangfeng and Li, Xin and Wu, Bo and Zhang, Jianwei and others},
  journal={arXiv preprint arXiv:2412.03603},
  year={2024}
}

@misc{veo,
    title={Veo},
    author={{Google DeepMind}},
    year={2025},
    url={https://deepmind.google/models/veo},
}

@misc{kling,
    author={{Kuaishou}},
    title={Kling Video Model},
    year={2024},
    howpublished={\url{https://kling.kuaishou.com/en}},
}

@inproceedings{lin2024diffbir,
  title={Diffbir: Toward blind image restoration with generative diffusion prior},
  author={Lin, Xinqi and He, Jingwen and Chen, Ziyan and Lyu, Zhaoyang and Dai, Bo and Yu, Fanghua and Qiao, Yu and Ouyang, Wanli and Dong, Chao},
  booktitle={European conference on computer vision},
  pages={430--448},
  year={2024},
  organization={Springer}
}

@inproceedings{AirNet,
author = {Li, Boyun and Liu, Xiao and Hu, Peng and Wu, Zhongqin and Lv, Jiancheng and Peng, Xi},
title = {{All-In-One Image Restoration for Unknown Corruption}},
booktitle = {IEEE Conference on Computer Vision and Pattern Recognition},
year = {2022},
address = {New Orleans, LA},
month = jun
}

@article{ma2023prores,
  title={Prores: Exploring degradation-aware visual prompt for universal image restoration},
  author={Ma, Jiaqi and Cheng, Tianheng and Wang, Guoli and Zhang, Qian and Wang, Xinggang and Zhang, Lefei},
  journal={arXiv preprint arXiv:2306.13653},
  year={2023}
}

@article{zhang2025perceive,
  title={Perceive-ir: Learning to perceive degradation better for all-in-one image restoration},
  author={Zhang, Xu and Ma, Jiaqi and Wang, Guoli and Zhang, Qian and Zhang, Huan and Zhang, Lefei},
  journal={IEEE Transactions on Image Processing},
  year={2025},
  publisher={IEEE}
}

@inproceedings{dit,
  title={Scalable diffusion models with transformers},
  author={Peebles, William and Xie, Saining},
  booktitle={Proceedings of the IEEE/CVF international conference on computer vision},
  pages={4195--4205},
  year={2023}
}

@misc{videoworldsimulators2024,
    title={Video generation models as world simulators},
    author={Tim Brooks and Bill Peebles and Connor Holmes and Will DePue and Yufei Guo and Li Jing and David Schnurr and Joe Taylor and Troy Luhman and Eric Luhman and Clarence Ng and Ricky Wang and Aditya Ramesh},
    year={2024},
    url={https://openai.com/research/video-generation-models-as-world-simulators},
}

@article{ho2022imagen,
  title={Imagen video: High definition video generation with diffusion models},
  author={Ho, Jonathan and Chan, William and Saharia, Chitwan and Whang, Jay and Gao, Ruiqi and Gritsenko, Alexey and Kingma, Diederik P and Poole, Ben and Norouzi, Mohammad and Fleet, David J and others},
  journal={arXiv preprint arXiv:2210.02303},
  year={2022}
}

@article{blattmann2023stable,
  title={Stable video diffusion: Scaling latent video diffusion models to large datasets},
  author={Blattmann, Andreas and Dockhorn, Tim and Kulal, Sumith and Mendelevitch, Daniel and Kilian, Maciej and Lorenz, Dominik and Levi, Yam and English, Zion and Voleti, Vikram and Letts, Adam and others},
  journal={arXiv preprint arXiv:2311.15127},
  year={2023}
}

@article{chen2023videocrafter1,
  title={Videocrafter1: Open diffusion models for high-quality video generation},
  author={Chen, Haoxin and Xia, Menghan and He, Yingqing and Zhang, Yong and Cun, Xiaodong and Yang, Shaoshu and Xing, Jinbo and Liu, Yaofang and Chen, Qifeng and Wang, Xintao and others},
  journal={arXiv preprint arXiv:2310.19512},
  year={2023}
}

@inproceedings{jin2023dnf,
  title={Dnf: Decouple and feedback network for seeing in the dark},
  author={Jin, Xin and Han, Ling-Hao and Li, Zhen and Guo, Chun-Le and Chai, Zhi and Li, Chongyi},
  booktitle={Proceedings of the IEEE/CVF conference on computer vision and pattern recognition},
  pages={18135--18144},
  year={2023}
}

@inproceedings{fang2022robust,
  title={A robust non-blind deblurring method using deep denoiser prior},
  author={Fang, Yingying and Zhang, Hao and Wong, Hok Shing and Zeng, Tieyong},
  booktitle={Proceedings of the IEEE/CVF conference on computer vision and pattern recognition},
  pages={735--744},
  year={2022}
}

@inproceedings{guo2022image,
  title={Image dehazing transformer with transmission-aware 3d position embedding},
  author={Guo, Chun-Le and Yan, Qixin and Anwar, Saeed and Cong, Runmin and Ren, Wenqi and Li, Chongyi},
  booktitle={Proceedings of the IEEE/CVF conference on computer vision and pattern recognition},
  pages={5812--5820},
  year={2022}
}

@inproceedings{liang2021swinir,
  title={Swinir: Image restoration using swin transformer},
  author={Liang, Jingyun and Cao, Jiezhang and Sun, Guolei and Zhang, Kai and Van Gool, Luc and Timofte, Radu},
  booktitle={Proceedings of the IEEE/CVF international conference on computer vision},
  pages={1833--1844},
  year={2021}
}

@article{deng2025video,
  title={Video Models Start to Solve Chess, Maze, Sudoku, Mental Rotation, and Raven'Matrices},
  author={Deng, Hokin},
  journal={arXiv preprint arXiv:2512.05969},
  year={2025}
}

@article{guo2025video,
  title={Are video models ready as zero-shot reasoners? an empirical study with the mme-cof benchmark},
  author={Guo, Ziyu and Chen, Xinyan and Zhang, Renrui and An, Ruichuan and Qi, Yu and Jiang, Dongzhi and Li, Xiangtai and Zhang, Manyuan and Li, Hongsheng and Heng, Pheng-Ann},
  journal={arXiv preprint arXiv:2510.26802},
  year={2025}
}

@article{luo2025v,
  title={V-ReasonBench: Toward Unified Reasoning Benchmark Suite for Video Generation Models},
  author={Luo, Yang and Zhao, Xuanlei and Lin, Baijiong and Zhu, Lingting and Tang, Liyao and Liu, Yuqi and Chen, Ying-Cong and Qian, Shengju and Wang, Xin and You, Yang},
  journal={arXiv preprint arXiv:2511.16668},
  year={2025}
}

@article{yang2025reasoning,
  title={Reasoning via Video: The First Evaluation of Video Models' Reasoning Abilities through Maze-Solving Tasks},
  author={Yang, Cheng and Wan, Haiyuan and Peng, Yiran and Cheng, Xin and Yu, Zhaoyang and Zhang, Jiayi and Yu, Junchi and Yu, Xinlei and Zheng, Xiawu and Zhou, Dongzhan and others},
  journal={arXiv preprint arXiv:2511.15065},
  year={2025}
}

@article{chen2025tivibench,
  title={TiViBench: Benchmarking Think-in-Video Reasoning for Video Generative Models},
  author={Chen, Harold Haodong and Lan, Disen and Shu, Wen-Jie and Liu, Qingyang and Wang, Zihan and Chen, Sirui and Cheng, Wenkai and Chen, Kanghao and Zhang, Hongfei and Zhang, Zixin and others},
  journal={arXiv preprint arXiv:2511.13704},
  year={2025}
}

@article{li2025viper,
  title={VIPER: Process-aware Evaluation for Generative Video Reasoning},
  author={Li, Yifan and Gu, Yukai and Min, Yingqian and Liu, Zikang and Du, Yifan and Zhou, Kun and Yang, Min and Zhao, Wayne Xin and Qiu, Minghui},
  journal={arXiv preprint arXiv:2512.24952},
  year={2025}
}

@article{tong2026cof,
  title={CoF-T2I: Video Models as Pure Visual Reasoners for Text-to-Image Generation},
  author={Tong, Chengzhuo and Chang, Mingkun and Zhang, Shenglong and Wang, Yuran and Liang, Cheng and Zhao, Zhizheng and An, Ruichuan and Zeng, Bohan and Shi, Yang and Dai, Yifan and others},
  journal={arXiv preprint arXiv:2601.10061},
  year={2026}
}

@misc{miniveo3reasoner,
    title = {MiniVeo3-Reasoner: Thinking with Videos from Open-Source Priors},
    author = {Wu, Jialong and Huang, Tianhao and He, Changjing and Long, Mingsheng},
    year = {2025},
    publisher = {GitHub},
    journal = {GitHub repository},
    howpublished = {\url{https://github.com/thuml/MiniVeo3-Reasoner}},
}

\clearpage
\appendix
\begin{center}
{\LARGE \textbf{Appendix for V-Bridge}}
\end{center}

\section{Details}\label{appendix:details}

\subsection{Training Details}\label{appendix:training-details}

\noindent\textbf{Data.} For the training data, we construct high-quality (HQ) and low-quality (LQ) image pairs using samples from FoundIR and RealCE. Specifically, FoundIR contains several common degradation categories, including blur, lowlight, JPEG compression, haze, rain, as well as combinations of multiple degradations. To maintain a balanced distribution, we perform uniform sampling across these degradation categories. In contrast, RealCE mainly consists of text degradation data in both Chinese and English, and its samples are treated as a single category during sampling.
Unless otherwise specified in the main paper, the default training set size used in our experiments is 1K. The detailed data construction procedure can be found in Sec.~\ref{sec:method:data}.

\begin{table}[htbp]
\centering
\caption{Training Configurations for V-Bridge.}
\label{tab:appendix:hyper-parameter}
\vspace{-8pt}
\begin{tabular}{l l}
\toprule
\textbf{Parameter} & \textbf{Value} \\
\midrule
\midrule
BF16 & True \\
Learning Rate & 2e-5 \\
LR Scheduler Type & constant\_with\_warmup \\
LR Warmup Steps & 100 \\
Adam Weight Decay & 3e-2 \\
Adam Epsilon & 1e-10 \\
Maximum Grad Norm & 0.05 \\
GPUs Per Node & 8 \\
Number of Nodes & 1 \\
Seed & 42 \\
\bottomrule
\end{tabular}
\vspace{-12pt}
\end{table}
\noindent\textbf{Details.} During training, the hyperparameters used in our experiments are summarized in Tab.~\ref{tab:appendix:hyper-parameter}. As described in Sec.~\ref{sec:method:training}, we adopt a three-stage training strategy. The primary difference across stages lies in the input video resolution, while all other training hyper-parameters remain consistent throughout the entire training process. Specifically, the three training stages use resolutions of 512, 720, and 960, respectively. The meaning of these resolutions is defined as follows. Taking 512 as an example, for each frame in the original video, we first resize the frame while preserving its aspect ratio such that the shorter side is scaled to 512. Then, a 512×512 patch is randomly cropped from the resized frame for training. The resolutions used in the other stages follow the same procedure.  The total training schedule consists of 300 epochs, which are evenly divided across the three stages.

\subsection{Test Details}\label{appendix:test-details}

\noindent\textbf{Benchmarks.} 
Next, we provide a detailed description of each benchmark used in our experiments, including the specific degradation types involved and their corresponding detailed information.

\begin{itemize}
    \item \textbf{FoundIR~\cite{foundir}.} This test set includes common degradation types such as blur, low-light, JPEG compression, haze, rain, as well as combinations of multiple degradations. It contains a total of 1,500 samples, with the majority of the data at resolutions of 1080p or higher.
    \item \textbf{Dense-Haze~\cite{dense-haze}.} Dense-Haze comprises 33 pairs of real-world hazy and corresponding haze-free images, capturing a variety of outdoor scenes. The hazy conditions are densely distributed and visually uniform, created using professional haze machines to introduce authentic atmospheric haze.
    \item \textbf{UHD-LL~\cite{uhd-ll}.} The first real UHD low-light image enhancement dataset, featuring 4K image pairs captured under diverse conditions with different levels of darkness and noise. The dataset provides a test set consisting of 115 image pairs for evaluation purposes.
    \item \textbf{UAV-Rain1K~\cite{uav-rain1k}.} This dataset is specifically designed for studying rain removal in aerial imagery captured by unmanned aerial vehicles (UAVs). It provides a comprehensive collection of images that reflect various rainy conditions and environmental scenarios. The test set consists of 220 paired images, each with an average resolution of around 1500 × 1000 pixels, offering sufficient detail for evaluating the performance of deraining algorithms.
    \item \textbf{HQ-NightRain~\cite{hq-nightrain}. } The test portion of the dataset consists of 300 image pairs and is categorized based on the type of nighttime rain degradation into three distinct subsets: rain streaks (RS), raindrops (RD), and mixed rain (SD), which combines both streaks and drops.
    \item \textbf{WeatherBench~\cite{weatherbench}.} This dataset contains images with snow, haze, and rain degradations. In our evaluation, we only utilize the snow subset, which provides 200 pairs of corresponding low-quality and high-quality images. It is worth noting that snow removal is an unseen task for our model, as this type of degradation is not included in the tasks encountered during training.
\end{itemize}

\noindent\textbf{Inference.} During inference, a unified prompt is employed across all tasks, with the full prompt details provided in Box.A.1 and A.2. Since generating high-resolution videos (e.g., 4K) directly with video generation models incurs substantial VAE decoding time, we adopt a practical approach for efficiency: frames with resolutions exceeding 2K are first resized to approximately 2K while maintaining their original aspect ratio for inference. The outputs are then rescaled back to the original resolution prior to metric computation. During generation, we use UniPC, adapted for flow matching, as the sampling scheduler with 50 sampling steps. The classifier-free guidance scale is set to 5.0, and the timestep shift is also set to 5.0.
\begin{center}
\begin{namedbox}[label=a.1]{Prompt for Image Restoration}\label{box:prompt}
A restoration-focused video strictly based on the input image. 
The camera is completely static with no movement, no zoom, and no rotation.
The original composition, objects, layout, and perspective are preserved exactly.
Focus on visual restoration and enhancement: remove noise, reduce blur, eliminate rain artifacts, 
    remove compression artifacts, and improve clarity, sharpness, and fine details while maintaining 
    natural textures, accurate colors, and balanced lighting. Only extremely subtle and natural temporal 
    consistency is allowed. The video should appear stable, clean, and realistic, as if the input image 
    has been gently restored over time.
\end{namedbox}
\end{center}
\begin{center}
\begin{namedbox}[label=a.2]{Negative Prompt for Image Restoration}\label{box:negative-prompt}
camera movement, panning, tilting, zooming, rotation, 
        scene change, object movement, new objects, object deformation, 
        style change, artistic style, illustration, painting, cartoon, 
        over-saturated colors, overexposure, underexposure, 
        motion blur, jitter, flickering, shaking, 
        low quality, worst quality, noise, blur, rain, fog, 
        compression artifacts, jpeg artifacts, aliasing, 
        text, subtitles, watermark, logo, 
        distorted anatomy, extra limbs, duplicated objects, 
        exaggerated motion, creative animation
\end{namedbox}
\end{center}

\noindent\textbf{Metrics.} In this work, we adopt \textbf{PSNR} and \textbf{SSIM} as evaluation metrics, where higher values indicate better performance. Following the computation methodology in FoundIR , these metrics are defined as:

\textbf{1. PSNR (Peak Signal-to-Noise Ratio):}
\begin{align}
\text{MSE} &= \frac{1}{HW} \sum_{i=1}^{H} \sum_{j=1}^{W} \big(I(i,j) - \hat{I}(i,j)\big)^2, \\
\text{PSNR} &= 10 \cdot \log_{10} \left( \frac{L^2}{\text{MSE}} \right)
\end{align}
where $I$ and $\hat{I}$ denote the ground-truth and reconstructed images, $H$ and $W$ are the height and width of the images, and $L$ is the maximum pixel value (e.g., 255 for 8-bit images).

\textbf{2. SSIM (Structural Similarity Index):}
\begin{align}
\text{SSIM}(I, \hat{I}) &= \frac{(2 \mu_I \mu_{\hat{I}} + C_1)(2 \sigma_{I\hat{I}} + C_2)}
{(\mu_I^2 + \mu_{\hat{I}}^2 + C_1)(\sigma_I^2 + \sigma_{\hat{I}}^2 + C_2)}
\end{align}
where $\mu_I, \mu_{\hat{I}}$ and $\sigma_I^2, \sigma_{\hat{I}}^2$ are the means and variances of $I$ and $\hat{I}$, $\sigma_{I\hat{I}}$ is their covariance, and $C_1, C_2$ are stabilizing constants.

Both metrics are computed per image and averaged over the dataset to obtain the final evaluation scores.

\section{Additional Experimental Results}


\subsection{Visualization Results}
\begin{figure*}[t]
\centering  
\includegraphics[width=0.9 \textwidth]{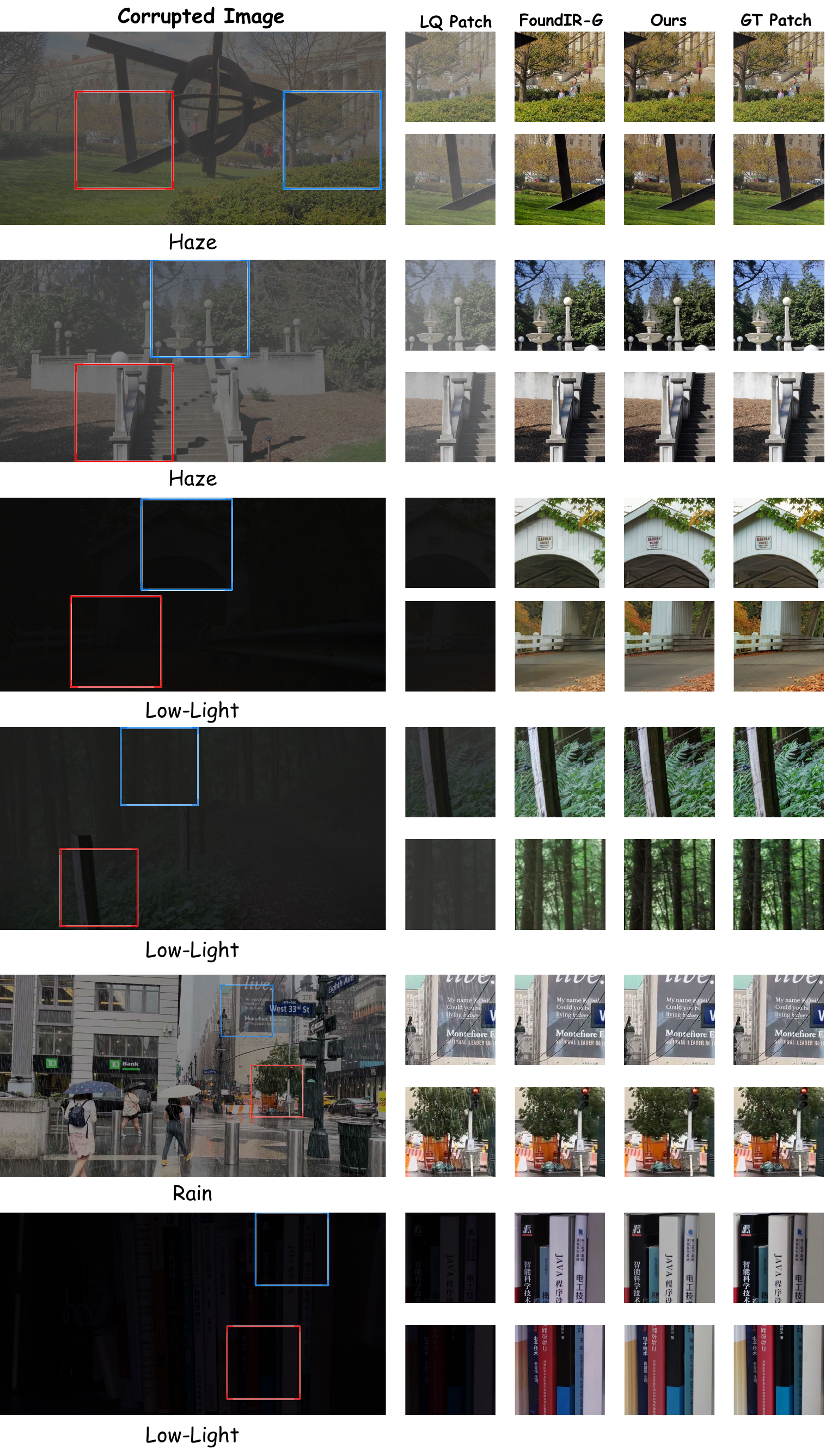} 
\vspace{-0.3cm}
\caption{Additional visualization results.}\label{fig:appendix:visual1}
\vspace{-5pt}
\end{figure*}

\begin{figure*}[t]
\centering  
\includegraphics[width=0.9 \textwidth]{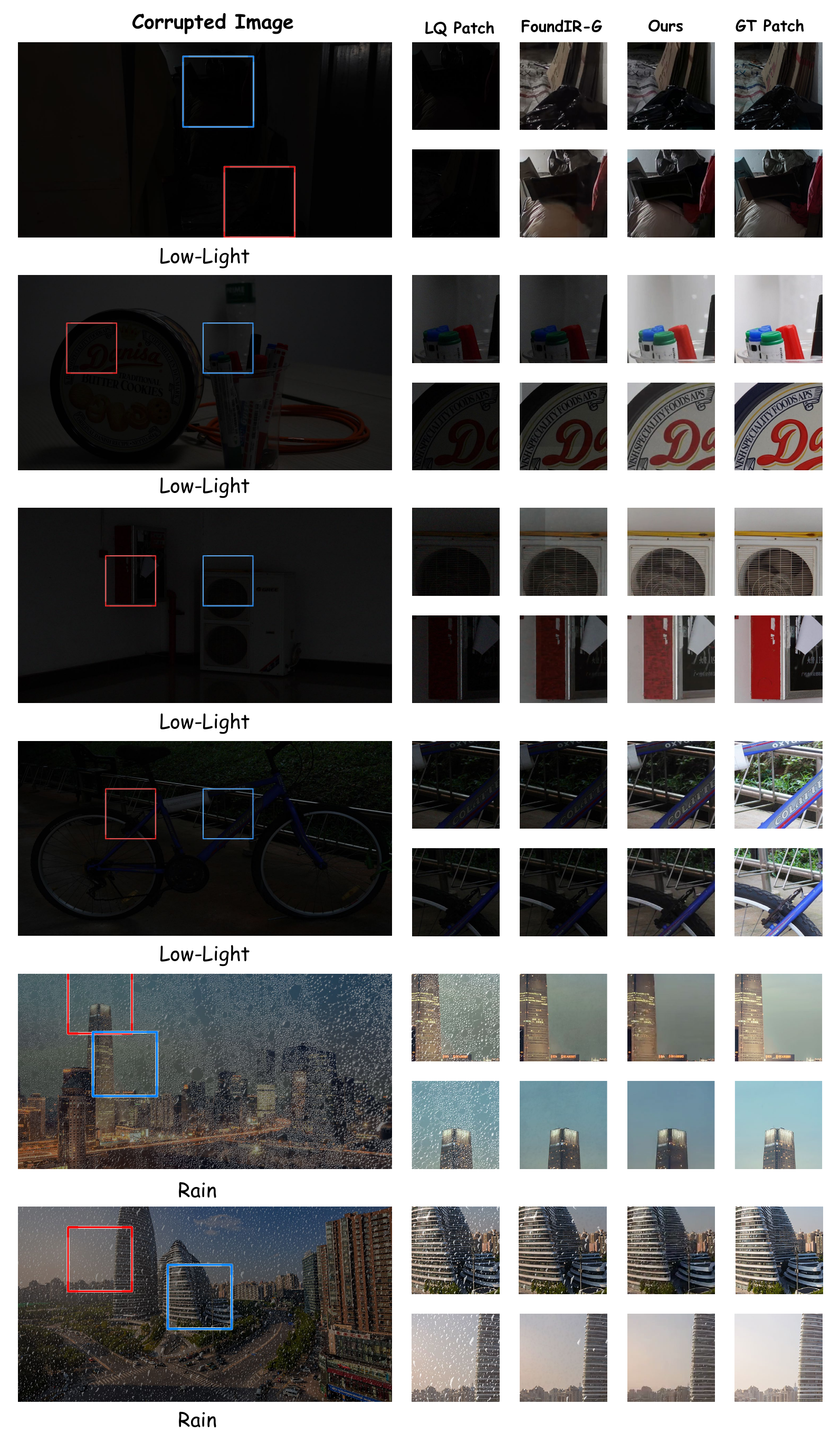} 
\vspace{-0.3cm}
\caption{Additional visualization results.}\label{fig:appendix:visual2}
\vspace{-5pt}
\end{figure*}

\begin{figure*}[t]
\centering  
\includegraphics[width=0.9 \textwidth]{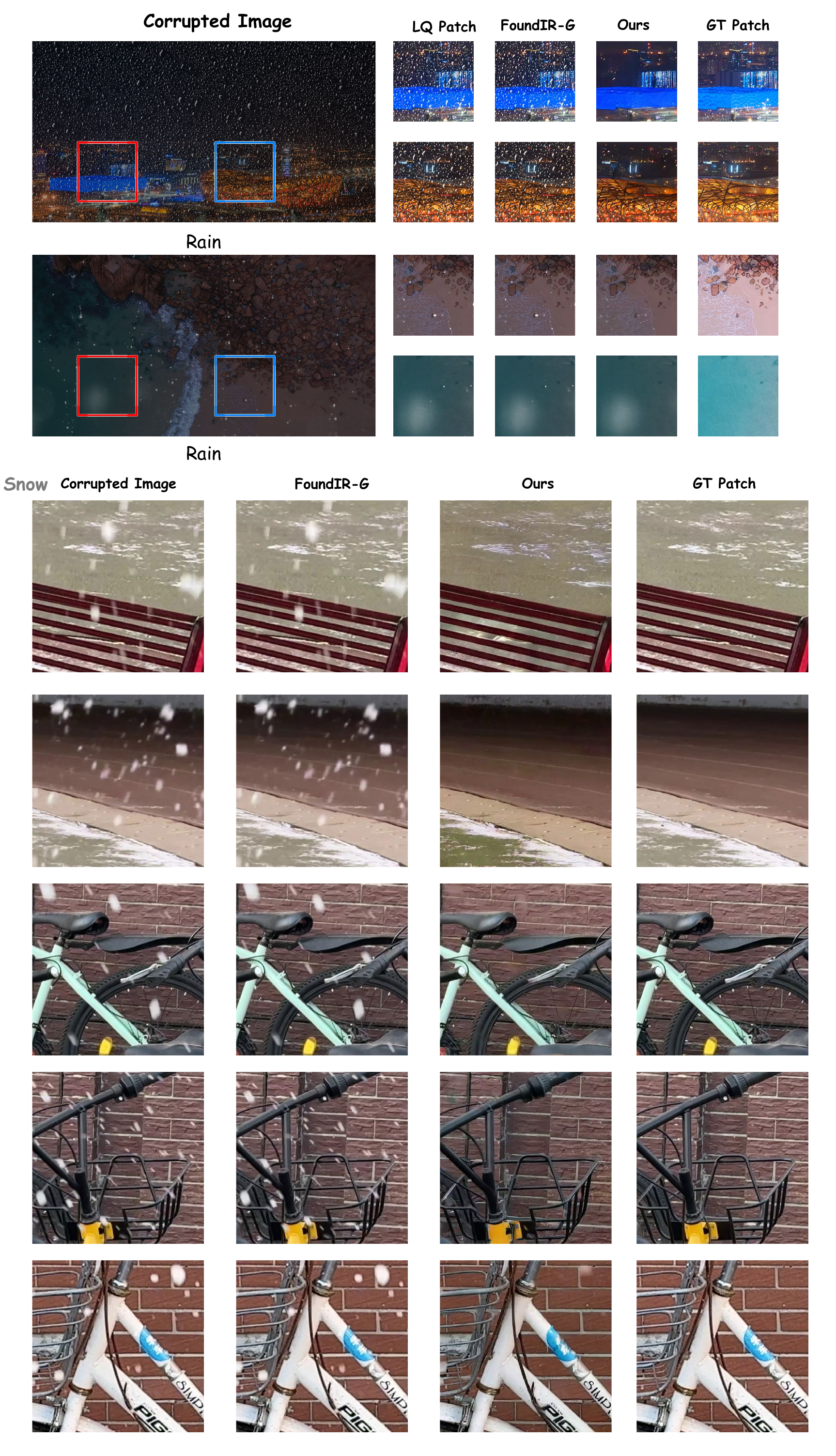} 
\vspace{-0.3cm}
\caption{Additional visualization results.}\label{fig:appendix:visual3}
\vspace{-5pt}
\end{figure*}

We present additional visualization results in Fig.\ref{fig:appendix:visual1}, Fig.\ref{fig:appendix:visual2}, and Fig.\ref{fig:appendix:visual3}.

\subsection{Failure Cases} 
\begin{figure*}[t]
\centering  
\includegraphics[width=0.9 \textwidth]{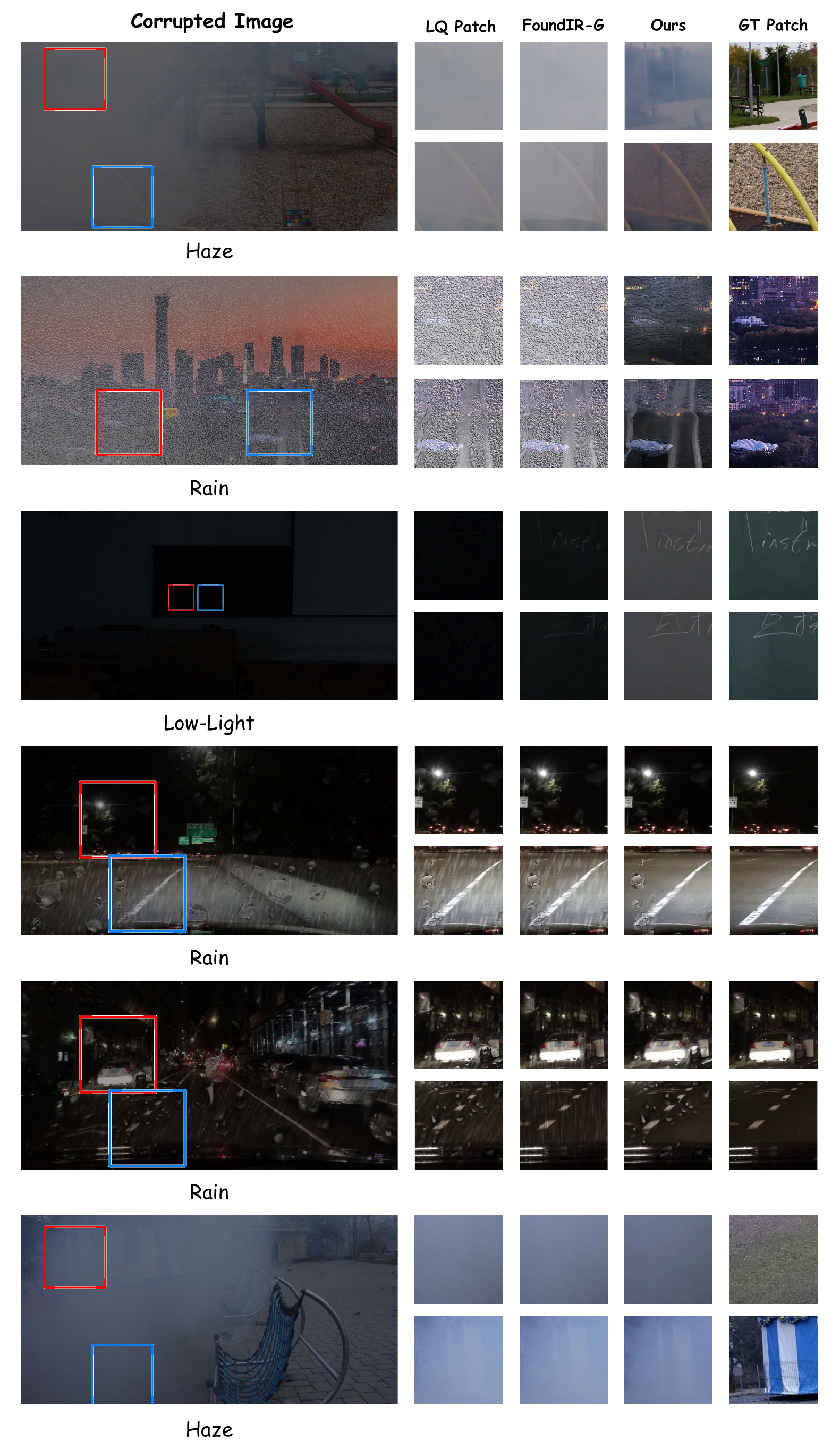} 
\vspace{-0.3cm}
\caption{Failure cases.}\label{fig:appendix:failure}
\vspace{-5pt}
\end{figure*}

We present several failure cases of our method as shown in Fig.\ref{fig:appendix:failure}, along with the corresponding visualization results of FoundIR-G on the same data.

\section{Limitations and Future Work}
This work primarily explores the feasibility of transferring the prior knowledge of video generation models to low-level vision tasks, and demonstrates the effectiveness and potential of such a transfer. However, the current approach faces two main limitations: task generalization and efficiency. In future work, we aim to extend the transfer of video generation priors to a broader range of tasks, further validating the potential of video generation models as foundation models for vision. Additionally, since the present study focuses solely on demonstrating potential, we did not incorporate any acceleration strategies to avoid confounding factors in the analysis. Moving forward, we plan to introduce acceleration techniques to make the approach practically usable in real-world scenarios.

\end{document}